\documentclass[article,jmlmc]{beg_32}      

\usepackage[hang]{footmisc}
\usepackage{xcolor}
\fancypagestyle{plain}{%
  \fancyhf{}
  \fancyfoot[R]{\small\bf\thepage }
  }
\renewcommand{\dmyy}{20}

\def\eqref#1{equation~(\ref{#1})}
\def\Eqref#1{Equation~(\ref{#1})}

\def\1{\bm{1}}

\def\rx{{\textnormal{x}}}

\def\rve{{\mathbf{e}}}

\def\rvx{{\mathbf{x}}}

\def\vtheta{{\bm{\theta}}}

\def\vepsilon{{\bm{\epsilon}}}

\def\vx{{\bm{x}}}

\def\mI{{\bm{I}}}

\DeclareMathAlphabet{\mathsfit}{\encodingdefault}{\sfdefault}{m}{sl}
\SetMathAlphabet{\mathsfit}{bold}{\encodingdefault}{\sfdefault}{bx}{n}

\newcommand{\R}{\mathbb{R}}

\usepackage{hyperref}       
\usepackage{nicefrac}       
\usepackage{booktabs}
\usepackage{mathtools}

\begin{document}

\volume{}
\title{Leveraging Local Variation in Data: Sampling and Weighting Schemes for Supervised Deep Learning}
\titlehead{Leveraging Local Variation of Data in Supervised Deep Learning}
\authorhead{P. Novello, G. Poette, D. Lugato, \& P.M. Congedo}
\corrauthor[1,2,3]{P. Novello}
\author[1]{G. Poëtte}
\author[1]{D. Lugato}
\author[2,3]{P.M. Congedo}
\corremail{paul.novello@outlook.fr}
\address[1]{CESTA, CEA, Le Barp, 33114, France}
\address[2]{CMAP, Ecole Polytechnique, 91120, Palaiseau, France}
\address[3]{Platon, Inria Paris Saclay, 91120, Palaiseau, France}

\dataO{mm/dd/yyyy}
\dataF{mm/dd/yyyy}

\abstract{In the context of supervised learning of a function by a neural network, we claim and empirically verify that the neural network yields better results when the distribution of the data set focuses on regions where the function to learn is steep. We first traduce this assumption in a mathematically workable way using Taylor expansion and emphasize a new training distribution based on the derivatives of the function to learn. Then, theoretical derivations allow constructing a methodology that we call Variance Based Samples Weighting (VBSW). VBSW uses labels local variance to weight the training points. This methodology is general, scalable, cost-effective, and significantly increases the performances of a large class of neural networks for various classification and regression tasks on image, text, and multivariate data.  We highlight its benefits with experiments involving neural networks from linear models to ResNet \cite{resnet} and Bert \cite{bert}.
}

\keywords{Supervised learning, importance weighting, learning theory, designs of experiments}

\maketitle

When a Machine Learning (ML) model is used to learn from data, the distribution of the training data set can have a substantial impact on its performance. More specifically, in Deep Learning (DL), several works have hinted at the importance of the training set. In  \cite{Bengio:2009:CL:1553374.1553380, teacherstudent}, the authors exploit the observation that a human will benefit more from easy examples than from harder ones at the beginning of a learning task. They construct a curriculum, inducing a change in the distribution of the training data set that makes a neural network achieve better results in an ML problem. With a different approach, Active Learning \cite{series/synthesis/2012Settles} modifies the distribution of the training data dynamically by selecting the data points that will make the training more efficient. Finally, in Reinforcement Learning, the distribution of experiments is crucial for the agent to learn efficiently. Moreover, the challenge of finding a good distribution is not specific to ML. Indeed, in the context of Monte Carlo estimation of a quantity of interest based on a random variable, Importance Sampling owes its efficiency to the construction of a second random variable, which is used instead to improve the estimation of this quantity. \cite{NIPS2010_3922} even make a connection between the success of likelihood ratio policy gradients and importance sampling, which shows that ML and Monte Carlo estimation, both distribution-based methods, are closely linked.

In this paper, we leverage the importance of the training set distribution to improve the performances of neural networks in supervised deep learning. We formalize supervised learning as a task which aims at  approximating a function $f$ with a model $f_{\vtheta}$ parametrized by $\vtheta$ using data points drawn from $X \sim d\mathbb{P}_X$, $X \in \mathcal{X}$. We build a new distribution $d\mathbb{P}_{\bar{X}}$ from the training points and their labels, based on the observation that $f_{\vtheta}$ needs more data points to approximate $f$ on the regions where it is steep. We derive an illustrative generalization bound involving the derivatives of $f$ that theoretically corroborates this observation. Therefore, we build $d\mathbb{P}_{\bar{X}}$ using Taylor expansion of the function $f$, which links the local behavior of $f$ to its derivatives.  

We first focus on the influence of using $d\mathbb{P}_{\bar{X}}$ instead of $d\mathbb{P}_X$ in simple approximation problems. To that end, we build a methodology for constructing and exploiting $d\mathbb{P}_{\bar{X}}$, that we call Taylor Based sampling (TBS). We then apply TBS to a more realistic problem based on the approximation of the solution of Bateman equations. Solving these equations is an important part of many numerical simulations of several phenomena (neutronic \citep{gaek,Dufek}, combustion \citep{Bisi}, detonic \citep{LUCOR}, computational biology \citep{transport-bio}, etc.). 

Then, we study the benefits of this approach for more general machine learning problems. In these cases, exploiting $d\mathbb{P}_{\bar{X}}$ is less straightforward. Indeed, we do not know the derivatives of $f$, and we cannot obtain labels for new data points sampled from this distribution. To tackle these problems, we show that variance is an approximation of Taylor expansion up to a certain order. Then we leverage the link between sampling and weighting to construct a methodology called Variance Based Sample Weighting (VBSW). This methodology weights each training data point using the local variance of their neighbor labels to simulate the new distribution. We specifically investigate its application in deep learning, where we apply VBSW within the feature space of a pre-trained neural network. We validate VBSW for deep learning by obtaining performance improvements on various tasks like classification and regression of text, from Glue benchmark \citep{glue}, image, from MNIST \citep{mnist} and Cifar10 \citep{cifar} and multivariate data, from UCI machine learning repository \footnote{\url{http://archive.ics.uci.edu/ml}}, for several models ranging from linear regression to Bert \citep{bert} or ResNet20 \citep{resnet}. We also conduct analyses on the complementarity of VBSW with other weighting techniques and its robustness to label noise.\\

\section{Related works}
\label{sec:related_vbsw}

This work introduces contributions that rely on different elements. First, many techniques aim to alter the training distribution to improve the prediction error of neural networks. Second, finding generalization bounds for neural networks is the goal of various works in machine learning research. Finally, the methodology of constructing a sampling distribution for statistical analysis is used for importance sampling and designs of experiments.\\

\textbf{Modified learning distributions.} Some works are dedicated to improving neural network performances by modifying the training distribution, either by weighting data points or by inducing sample selection. Active learning \citep{series/synthesis/2012Settles} adapts the training strategy to a learning problem by introducing an online data point selection rule. \cite{Gal} uses the variational properties of Bayesian neural network to design a rule that focuses the training on points that will reduce the prediction uncertainty of the neural network. In \cite{Konyushkova}, the construction of the selection rule is itself taken as a machine learning problem. See \cite{series/synthesis/2012Settles} for a review of more classical active learning methods. Unlike active learning, and similarly to VBSW, some other methods aim at introducing diverse \textit{a priori} evaluations of sample importance. While curriculum learning \citep{Bengio:2009:CL:1553374.1553380, teacherstudent} starts the training with easier examples, self-paced learning \citep{Kumar, SPCL} downscales harder examples. However, some works have proven that focusing on harder examples at the beginning of the learning could accelerate it: \cite{ShrivastavaGG16} performs hard example mining to give more importance to harder examples by selecting them primarily. This work also focuses on defining hard examples but does so with an original, mathematical way based on $f$ derivatives and local variance. It also stands out from the aforementioned techniques for how it modifies the distribution based on this information. Indeed, it suggests and justifies that a neural network should spend more learning time on subspaces of $\mathcal{X}$ which contain harder examples.\\

\textbf{Generalization bounds.} As an argument to motivate our approach, we derive a generalization bound. The construction of Generalization bounds for the learning theory of neural networks has motivated many works (see \cite{review} for a review). In \cite{VC,NTVC}, the authors focus on  Vapnik Chervonenkis (VC) dimension, a measure that depends on the number of parameters of neural networks. \cite{contraction} introduces a compression approach that aims at reducing the number of model parameters to investigate its generalization capacities. Probably Approximately Correct (PAC) Bayes analysis constructs generalization bounds using \textit{a priori} and \textit{a posteriori} distributions over the possible models. It is investigated, for example, in \cite{neyshabur2018a, barlett}. \cite{exploring, Xu2012} links PAC-Bayes theory to the notion of sharpness of a neural network, i.e. its robustness to small perturbation. While previous works often mention the sharpness of the model, our bound includes the derivatives of $f$, which can be seen as an indicator of the sharpness of the function to learn. Even if it uses elements of previous works, like the Lipschitz constant of $f_{\vtheta}$, our work does not pretend to tighten and improve the already existing generalization bounds. It only emphasizes the intuition that the neural network would need more points to capture sharper functions. In a sense, it investigates the robustness to perturbations in the input space, not in the parameter space. \\

\textbf{Examples weighting.} VBSW can be categorized as an examples weighting, or importance weighting algorithm. The idea of weighting the data set has already been explored in different ways and for various purposes. Examples weighting is used in \cite{imbalance} to tackle the class imbalance problem by weighting rarer, so harder examples. On the contrary, in \cite{noisy} it is used to solve the noisy label problem by focusing on cleaner, so easier examples. All these ideas show that depending on the application, examples weighting can be performed in an opposed manner. Some works aim at going beyond this opposition by proposing more general methodologies. In \cite{Chang}, the authors use the variance of the prediction of each point throughout the training to decide whether it should be weighted or not.  A meta-learning approach is proposed in \cite{Mengye}, where the authors choose the weights after an optimization loop included in the training. VBSW stands out from the previously mentioned examples weighting methods because it does not aim at solving dataset-specific problems like class imbalance or noisy labels. It is built on a more general assumption that a model would simply need more points to learn more complicated functions. The resulting weighting scheme verifies recent findings of \cite{understandingIW} where authors conclude that in classification, a good set of weights would put importance on points close to the decision boundary.\\

\textbf{Importance sampling.} The challenge of finding a good distribution is not specific to machine learning. Indeed, in the context of Monte Carlo estimation of a quantity of interest based on a random variable, importance sampling owes its efficiency to the construction of a second random variable, which is used instead to improve the estimation of this quantity. \cite{NIPS2010_3922} even make a connection between the success of likelihood ratio policy gradients and importance sampling, which shows that machine learning and Monte Carlo estimation, both distribution-based methods, are closely linked. Moreover, some previously mentioned methods use importance sampling to design the weights of the data set or to correct the bias induced by the sample selection \citep{Katharopoulos2018NotAS}. In this work, we construct a new distribution that could be interpreted as an importance distribution. However, we weigh the data points to simulate this distribution. It does not aim at correcting a bias induced by this distribution.\\

\textbf{Designs of experiments.} Some methodologies are dedicated to the construction of data sets in the context of statistical analysis. These methodologies are called designs of experiments. In our case, the construction of a new training distribution could be seen as a design of experiments for learning. However, popular designs of experiments used for regression are either space-filling designs or model-based designs. Space-filling designs, like Latin hypercube sampling \citep{lhs} or maximin designs \citep{minimax}, aims at spreading the learning points to cover the input space as much as possible. Model-based designs use characteristics of $f_{\vtheta}$ to adapt the training distribution. Such designs can look to maximize the entropy of the prediction \citep{maxent} or minimize its uncertainty \citep{ego}. These last designs of experiments can be conducted sequentially, getting close to active learning \citep{alm, almnn, alc}. Our methodology does not depend on $f_{\vtheta}$, nor aims at filling the input space. Instead, its goal is to adapt the design of experiments to characteristics of $f$ in order to reduce the prediction error.\\

\section{Link between local variations and learning}

Let us first remind some basics on supervised machine learning. We formalize the supervised machine learning task as approximating a function $f: \mathbf{S} \subset \mathbb{R}^{n_i} \rightarrow \mathbb{R}^{n_o}$ with a machine learning model $f_{\vtheta}$ parametrized by $\vtheta$, where $\mathbf{S}$ is a measured sub-space of $\mathbb{R}^{n_i}$ depending on the application. To this end, we are given a training data set of $N$ points, $\{\vx_1, ... , \vx_N\} \in \mathbf{S} $, drawn from $\rvx\sim d\mathbb{P}_{\rvx}$ and their point-wise values, or labels $\{f(\vx_1),...,f(\vx_N)\}$. Parameters $\vtheta$ have to be found in order to minimize an integrated loss function $J_{\rvx}(\vtheta) = \mathbb{E} [L(f_{\vtheta}(\vx), f(\vx))]$, with $L$ the loss function, $L: \mathbb{R}^{n_o} \times \mathbb{R}^{n_o}  \rightarrow \mathbb{R}$. The data allow estimating $J_{\rvx}(\vtheta)$ by $\widehat{J_{\rvx}}(\vtheta) =  \sum_{i=1}^N \omega_i  L(f_{\vtheta}(\vx_i), f(\vx_i))$, with $\{\omega_1,...,\omega_N\} \in \R$ estimation weights, generally equal to $\frac{1}{N}$. Then, an optimization algorithm is used to find a minimum of $\widehat{J_{\rvx}}(\vtheta)$  w.r.t. $\vtheta$. 

\subsection{Illustration of the link using derivatives}
\label{sec:gb}

\begin{figure}[!h]
  \begin{center}
      \includegraphics[trim=0.5cm 1cm 0.5cm 0.5cm,width=0.9\linewidth]{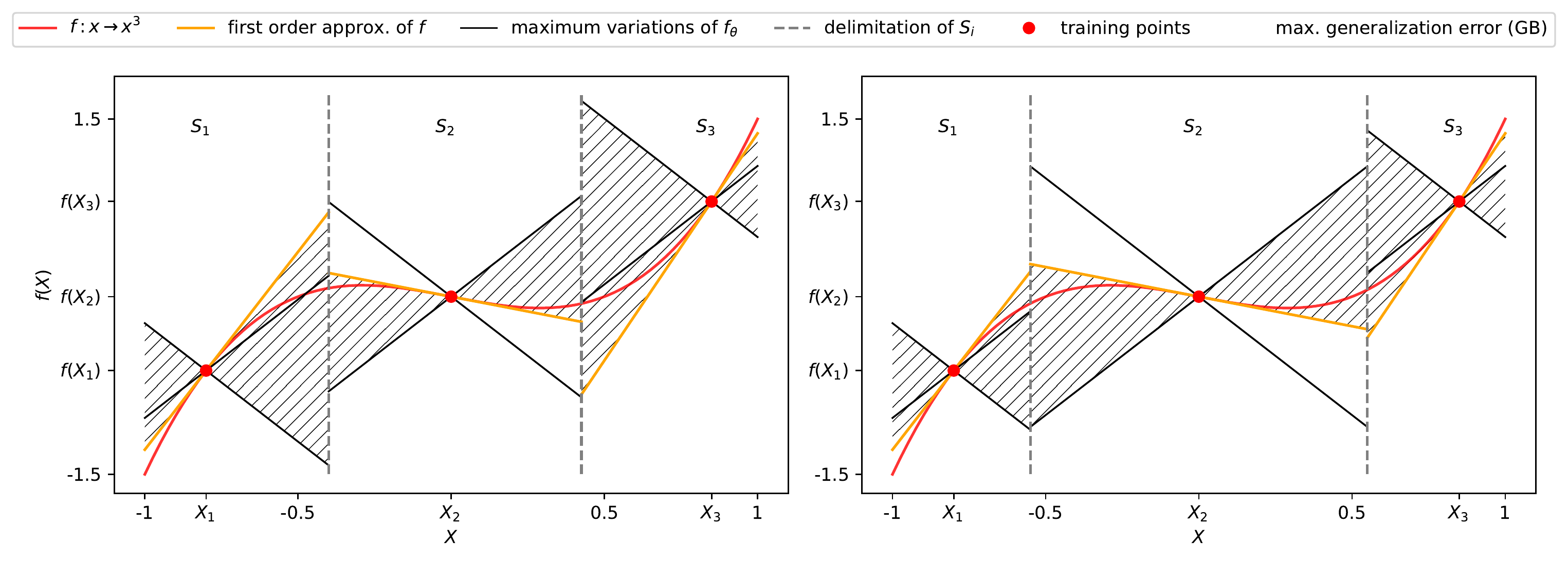}
  \end{center}
  \caption{\small Illustration of the GB. The maximum error (the GB), at order $\mathcal{O}(|S_i|^4)$, is obtained by comparing the maximum variations of $f_{\vtheta}$, and the first order approximation of $f$, whose trends are given by $K_ {\vtheta}$ and $f'(\vx_i)$. We understand visually that because $f'(\vx_1)$ and $f'(\vx_3)$ are higher than $f'(\vx_2)$, the GB is improved more efficiently by reducing $S_1$ and $S_3$ than $S_2$. \label{fig:gb}}
  
\end{figure}

In the following, we illustrate the intuition with a Generalization Bound (GB) that include the derivatives of $f$, provided that these derivatives exist. The goal of the approximation problem is to be able to generalize to points not seen during the training. The generalization error $\mathcal{J}_{\rvx}(\vtheta)= J_{\rvx}(\vtheta)  - \widehat{J_{\rvx}}(\vtheta)$ thus needs to be as small as possible. Let $S_i$, $i \in \{1,...,N\}$ be some sub-spaces of $\mathbf{S}$ such that $\mathbf{S} = \bigcup_{i=1}^N S_i$, $ \bigcap_{i=1}^N S_i =$ \O, and $\vx_i \in S_i$. Suppose that $L$ is the squared $L_2$ error, $n_i=n_o=1$, $f$ is differentiable, $f_{\vtheta}$ is $K_{\vtheta}$-Lipschitz and satisfies the conditions of Hornik theorem \cite{hornik}. Provided that $|S_i| < 1$, we show that

\begin{equation}\label{gb}
    \mathcal{J}_{\rx}(\vtheta) \leq \sum_{i=1}^N   (|f'(x_i)| +  K_{\vtheta})^2\frac{|S_i|^3}{4} +    \mathcal{O}(|S_i|^4),
\end{equation}

where $|S_i|$ is the volume of $S_i$ ($|S_i| = \int_{S_i} d \mathbb{P}_{\rvx}$). The proof can be found in \textbf{\hyperref[appA]{Appendix A}}. We see that in the regions where $f'(\vx_i)$ is high, quantity $|S_i|$ has a stronger impact on the GB. This idea is illustrated in Figure \ref{fig:gb}, which visually shows that the generalization bound increases  when $|S_i|$ and $f'(\vx_i)$ are high at the same time for approximating the function $f:x \rightarrow x^3$. Since $|S_i|$ can be seen as a metric for how close data points are around $\vx_i$ (the smaller $|S_i|$ is, the closer $\vx_i$ is to its neighbors), the GB can be reduced more efficiently by adding more points around $\vx_i$ in these regions. This bound also involves $K_{\vtheta}$, the Lipschitz constant of the neural network, which has the same impact as $f'(\vx_i)$. It also illustrates the link between the Lipschitz constant and the generalization error, which has been pointed out by several works like \cite{Gouk}, \cite{barlett} and \cite{qian2018lnonexpansive}.

\subsection{A sampling scheme based on Taylor Approximation}
\label{sec:tbs}

\Eqref{gb} formalizes a link between generalization error and derivatives of $f$. These derivatives are expressed at order $n=1$ for analytical reasons, but in this work we explore the use of derivatives of order $n>1$. Using Taylor expansion at order $n$ on $f$ and supposing that $f$ is $n$ times differentiable:

\begin{equation*}
    f(\vx+\vepsilon) \underset{\mathrm{\|\vepsilon \| \rightarrow 0}}{=} \sum_{0 \leq |\boldsymbol{k}| \leq n}  \boldsymbol{{\epsilon}^k} \frac{\partial^{\boldsymbol{k}}f(\vx)}{\boldsymbol{k}!} + \mathcal{O}(\boldsymbol{\epsilon^{n}}). 
    \label{taylor}
\end{equation*}

The quantity $f(\vx+ \vepsilon) - f(\vx) = \sum_{1 \leq |\boldsymbol{k}| \leq n}  \boldsymbol{{\epsilon}^k} \frac{\partial^{\boldsymbol{k}}f(\vx)}{\boldsymbol{k}!} + \mathcal{O}(\boldsymbol{\epsilon^{n}})$ gives an indication on how much $f$ changes around $\vx$. By neglecting the orders above $\boldsymbol{\epsilon^{n}}$, it is then possible to find the regions of interest by focusing on $Df^{n}_{\vepsilon}$, defined as:

\begin{equation}
    Df^{n}_{\vepsilon}(\vx) =  \sum_{1 \leq |\boldsymbol{k}| \leq n} \boldsymbol{{\epsilon}} ^k \frac{   (\partial^{\boldsymbol{k}}f(\vx))^2}{\boldsymbol{k}!},
    \label{dn}
\end{equation}

Where $\boldsymbol{k}$ is a multi-index, i.e. $\boldsymbol{k}=(k_1,...,k_{n_i})$ is a vector of $n_i$ non negative integers, $|\boldsymbol{k}| = \sum_{i=1}^{n_i} k_i$, $\boldsymbol{k}! = k_1!\times ... \times k_{n_i}!$, $ \boldsymbol{{\epsilon}^k} = \epsilon_1^{k_1}\times ... \times \epsilon_{n_i}^{k_{n_i}}$, $\partial^{\boldsymbol{k}} = \frac{\partial^{k_1}}{\partial x_1^{k_1}} \times ... \times \frac{\partial^{k_{n_i}}}{\partial x_{n_i}^{k_{n_i}}} $. Note that $Df^{n}_{\vepsilon}$ is evaluated using $(\partial^{\boldsymbol{k}}f(\vx))^2$ instead of $\partial^{\boldsymbol{k}}f(\vx)$ for derivatives not to cancel each other. To avoid these cancellations, the absolute could have been used, but we will see in Lemma \ref{estim} that the square value ensures interesting asymptotical properties. $f$ will be steeper and more irregular in the regions where $\vx \rightarrow Df^{n}_{\vepsilon}(\vx)$ is higher. To focus the training set on these regions, one can use $\{Df^{n}_{\vepsilon}(\vx_1),...,Df^{n}_{\vepsilon}(\vx_N)\}$ to construct a probability density function (pdf) and sample new data points from it. 

In this part, we empirically verify that using Taylor expansion to construct a new training distribution has a beneficial impact on the performances of a neural network. To this end, we construct a methodology, that we call Taylor Based Sampling (TBS), that generates a new training data set based on the metric \eqref{dn}. To focus the training set on the regions of interest, i.e. regions of high $\{Df^{n}_{\vepsilon}(\vx_1),...,Df^{n}_{\vepsilon}(\vx_N)\}$, we use this metric to construct a probability density function (pdf) - which is possible since $Df^{n}_{\vepsilon}(\vx) \geq 0$ for all $\vx \in \mathbf{S}$. It remains to normalize it but in practice it is enough considering a distribution $d\mathbb{P}_{\bar{\rvx}} \propto Df^{n}_{\vepsilon}$. Here, to approximate $d\mathbb{P}_{\bar{\rvx}}$ we use a Gaussian Mixture Model (GMM) with pdf $d\mathbb{P}_{\bar{\rvx}, GMM}$ that we fit to $\{Df^{n}_{\vepsilon}(\vx_1),...,Df^{n}_{\vepsilon}(\vx_N)\}$ using the Expectation-Maximization (EM) algorithm.  $N'$ new data points $\{\bar{\vx}_1,...,\bar{\vx}_{N'}\}$, can be sampled, with $\bar{\rvx} \sim d\mathbb{P}_{\bar{\rvx}, GMM}$. Finally, we obtain $\{f(\bar{\vx}_1),...,f(\bar{\vx}_{N'})\}$, add it to $\{f(\vx_1),...,f(\vx_N)\}$ and train our neural network on the whole data set.

TBS is described in Algorithm \ref{TBS}. \textbf{Line 1:} The parameter $\vepsilon$, the number of Gaussian distribution $n_{\text{GMM}}$ and $N'$ is chosen in order to avoid sparsity of $\{\bar{\vx}_{1},...,\bar{\vx}_{N'}\}$ over $\mathbf{S}$. \textbf{Line 2:} Without \textit{a priori} information on $f$, we sample the first points uniformly in a subspace $\mathbf{S}$. \textbf{Line 3-7:} We construct $\{Df^{n}_{\vepsilon}(\vx_1),...,Df^{n}_{\vepsilon}(\vx_N)\}$, and then $d\mathbb{P}_{\bar{\rvx}, GMM}$ to be able to sample points accordingly. \textbf{Line 8:} Because the support of a GMM is not bounded, some points can be sampled outside $\mathbf{S}$. We discard these points and sample until all points are inside $\mathbf{S}$. This rejection method is equivalent to sampling points from a truncated GMM. \textbf{Line 9-10:} We construct the labels and add the new points to the initial data set.

\begin{algorithm}
   \caption{Taylor Based Sampling (TBS)}
   \label{TBS}

    {\bfseries Inputs: } $\vepsilon$, $N$, $N'$, $n_{\text{GMM}}$, $n$\;
    Sample $\{\vx_1, ... , \vx_N\}$ from $\rvx \sim \mathcal{U}(\mathbf{S})$\;
   \For{$0 \leq k \leq n$}{
    Compute $\{\partial^{\boldsymbol{k}}f(\vx_1),...,\partial^{\boldsymbol{k}}f(\vx_N)\}$\;
   }
    Compute $\{Df^{n}_{\vepsilon}(\vx_1),...,Df^{n}_{\vepsilon}(\vx_N)\}$ using \eqref{dn}\;
    Approximate $d\mathbb{P}_{\bar{\rvx}} \propto Df^{n}_{\vepsilon}$ with a GMM using EM algorithm to obtain a density $d\mathbb{P}_{\bar{\rvx}, GMM}$\;
    Sample $\{\bar{\vx}_{1},...,\bar{\vx}_{N'}\}$ using rejection method to sample inside $\mathbf{S}$\;
    Compute $\{f(\bar{\vx}_{1}),...,f(\bar{\vx}_{N'})\}$\;
    Add $\{f(\bar{\vx}_{1}),...,f(\bar{\vx}_{N'})\}$ to $\{f(\vx_{1}),...,f(\vx_N)\}$\;

\end{algorithm}

\subsection{Taylor based sampling}
\subsubsection{Application to simple functions}

\begin{figure}[!h]
   \begin{subfigure}{0.49\textwidth}
     \includegraphics[width=1.0\linewidth]{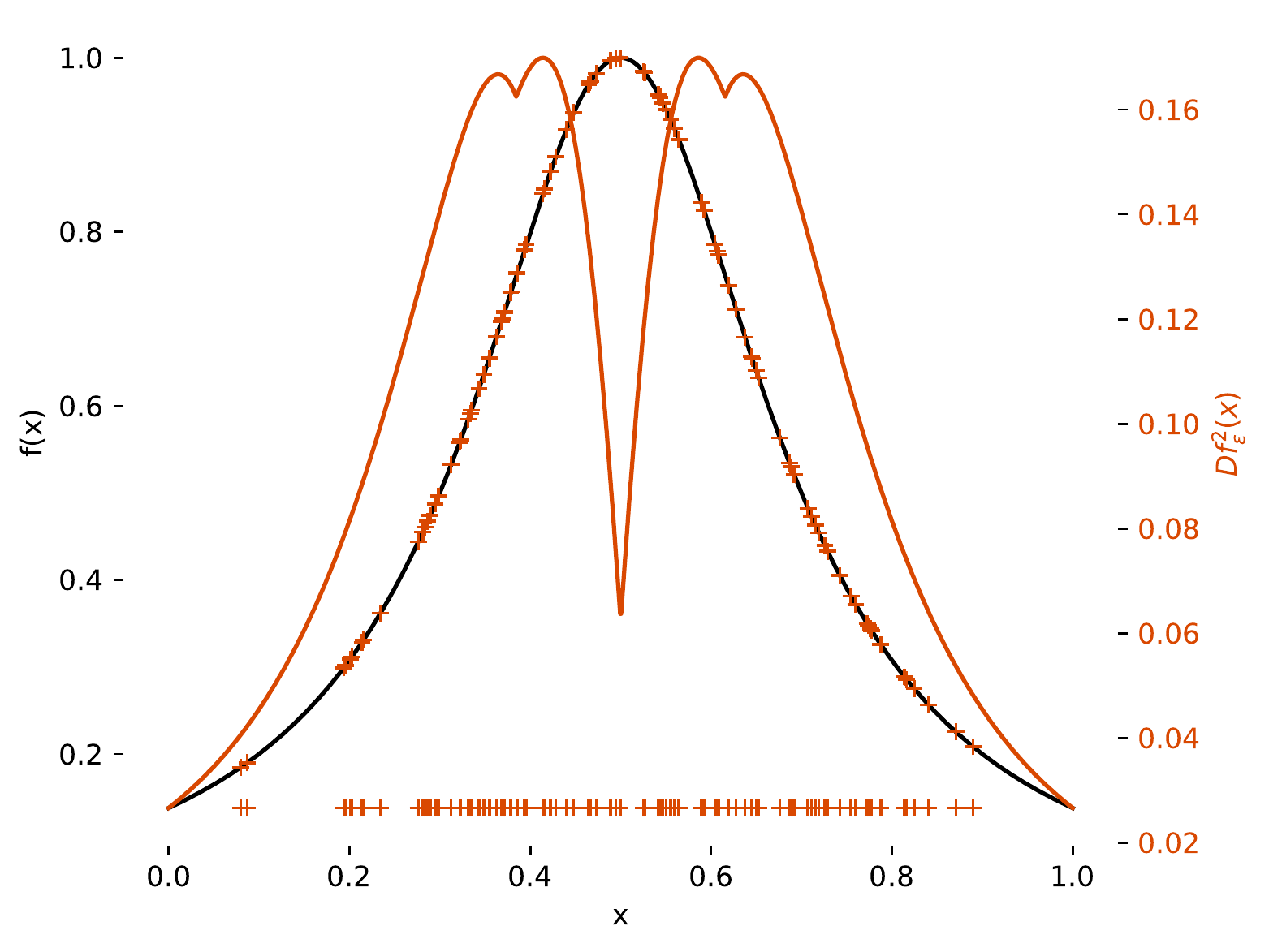}
     
   \end{subfigure}
   \begin{subfigure}{0.49\textwidth}
     \includegraphics[width=1.0\linewidth]{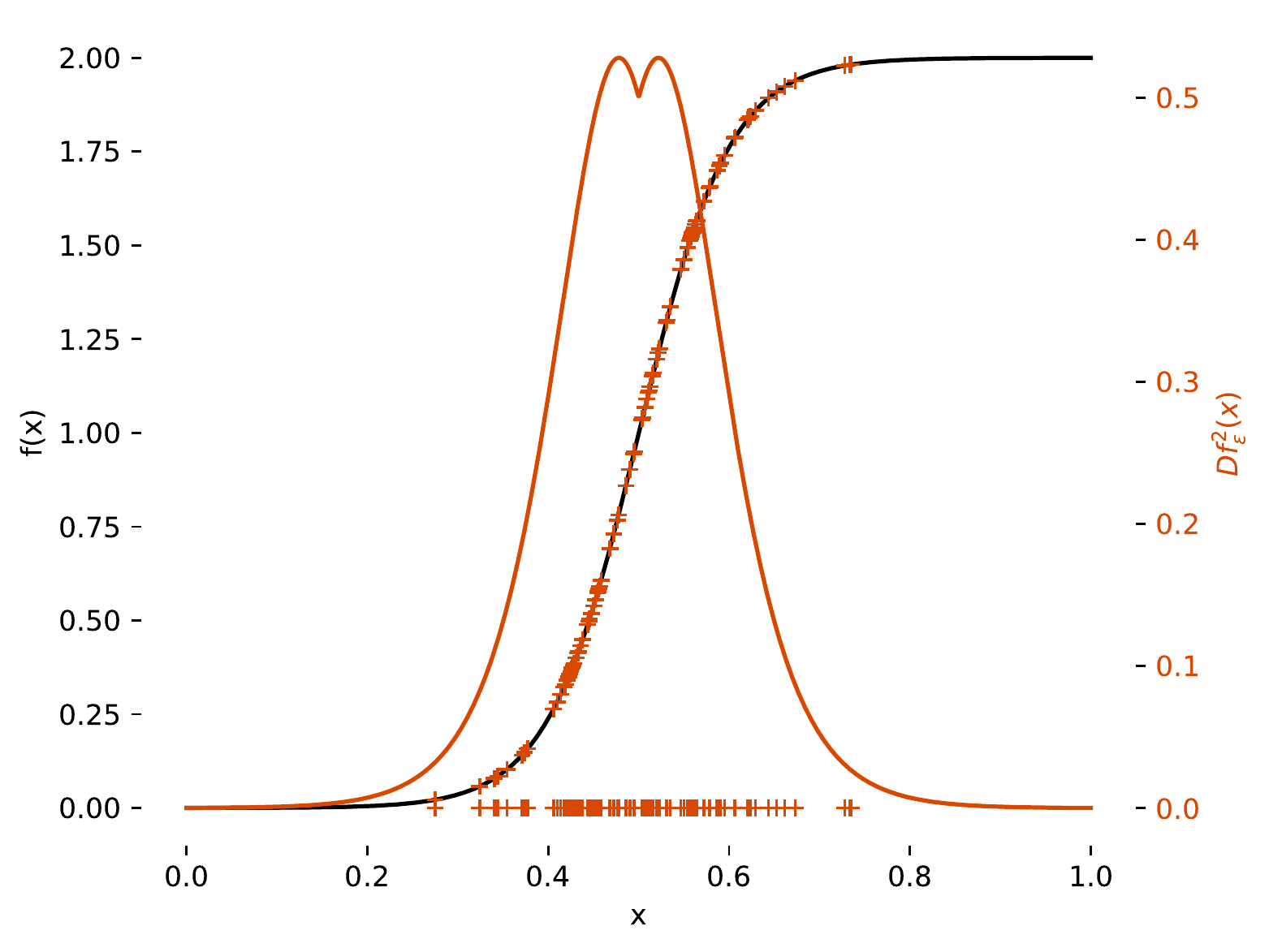}
     
   \end{subfigure}
   \caption{\textbf{Left:} (left axis) Runge function w.r.t x and (right axis) $x \rightarrow Df^{2}_{\vepsilon}(x)$. Points sampled using TBS are plotted on the x-axis and projected on $f$. \textbf{Right:} Same as left, with hyperbolic tangent function.}
   \label{f1}
\end{figure}

To illustrate the benefits of TBS compared to a uniform, basic sampling (BS), we apply it to two simple functions: hyperbolic tangent and Runge function. We chose these functions because they are differentiable and have a clear distinction between flat and steep regions. These functions are displayed in Figure \ref{f1}, as well as the map  $\vx \rightarrow Df^{2}_{\vepsilon}(\vx)$.

All neural networks have been implemented in \texttt{Python}, with \texttt{Tensorflow} \citep{Tensorflow}. We use the Python package \texttt{scikit-learn} \citep{scikit-learn} to construct $d\mathbb{P}_{\bar{\rvx}, GMM}$. The network chosen for this experiment is a Multi Layer Perceptron (MLP) with one layer of $8$ neurons and relu activation function, that we trained alternatively with BS and TBS using Adam optimizer \cite{adam} with the defaults tensorflow implementation hyperparameters, and Mean Squared Error loss function. We first sample $\{\vx_1, ... , \vx_N\}$ according to a regular grid. To compare the two methods, we add $N'$ additional points sampled using BS to create the BS data set, and then $N'$ other points sampled with TBS to construct the TBS data set. As a result, each data set have the same number of points $(N + N')$. We repeated the method for several values of $n$, $n_{\text{GMM}}$ and $\vepsilon$, to fine tune these parameters and finally selected $n=2$, $n_{\text{GMM}}=3$ and $\vepsilon = 10^{-3}$.

\begin{table}
  \centering
  \begin{tabular}{ccc}
  
   Sampling & $L_2$ error  &  $L_{\infty}$ error  
   \\ \hline 
    & \multicolumn{2}{c}{$f$: Runge $(\times 10^{-2})$} \\ 
   
  BS & $1.45 \pm 0.62 $ & $5.31 \pm 0.86$   \\
  \textbf{TBS}     & $\boldsymbol{1.13} \pm 0.73 $   & $\boldsymbol{3.87} \pm 0.48 $    \\ \hline 
    & \multicolumn{2}{c}{$f$: tanh $ (\times 10^{-1})$} \\ 
    
  BS & $1.39 \pm 0.67 $ & $2.75 \pm 0.78$   \\
  \textbf{TBS}     & $\boldsymbol{0.95} \pm 0.50 $   & $\boldsymbol{2.25} \pm 0.61 $    \\
  \end{tabular}
  \caption{\label{sample-table} Comparison between BS and TBS. The metrics used are the $L_2$ and $L_{\infty}$ errors, displayed with a $95\%$ confidence interval. }
  \end{table}

Table \ref{sample-table} summarizes the $L_2$ and the $L_{\infty}$ norm of the error of $f_{\vtheta}$, obtained at the end of the training phase for $N+N' = 16$, with $N=8$. Those norms are estimated using the same test data set of $1000$ points. The values are the means of the $40$ independent experiments displayed with a $95\%$ confidence interval. These results illustrate the benefits of TBS over BS. Table \ref{sample-table} shows that TBS does not significantly improve $L_2$ error, but does so for $L_{\infty}$ error, which may explain the good results of VBSW for classification that we describe in Section \ref{sec:exp_vbsw}. Indeed, the accuracy will not be very sensitive to small output variations for a classification task since the output is rounded to 0 or 1. However, a high error increases the risk of misclassification, which can be limited by the reduction of $L_{\infty}$.

\subsubsection{Application to an ODE system}

We apply TBS to a more realistic case: the approximation of the resolution of the Bateman equations, an ODE system. In this system, $u$ is the velocity of the reacting particles. Depending on the physical field of interest, $u$ may be distributed according to a Maxwellian distribution (dense gas with chemical reactions for example) or may be distributed according to a distribution computed by another part of the code (this is the case in general for neutronic reactions or collisions in a rarefied plasma). 
\setlength\abovedisplayskip{0pt}
\setlength\belowdisplayskip{0pt}

\begin{equation*}
\begin{dcases}
    \partial_t u(t)   &= v\boldsymbol{\sigma_a} \cdot \boldsymbol{\eta}(t)u(t), \\
    \partial_t\boldsymbol{\eta}(t) &= v\boldsymbol{\Sigma_r} \cdot  \boldsymbol{\eta}(t)u(t),  \\
\end{dcases}
\text{,    with initial conditions    }
\begin{dcases}
    u(0) = u_0, \\
    \boldsymbol{\eta}(0) = \boldsymbol{\eta_0}. \\ 
\end{dcases},
\end{equation*}

with $u \in \mathbb{R}^+, \boldsymbol{\eta}\in(\mathbb{R}^{+})^M,\boldsymbol{\sigma}_a^T\in\mathbb{R}^M,\boldsymbol{\Sigma}_r \in\mathbb{R}^{M\times M}$. Here, $f: (u_0, \boldsymbol{\eta_0}, t) \rightarrow (u(t), \boldsymbol{\eta}(t))$. For physical applications, $M$ ranges from tens to thousands, but we consider the particular case $M=1$ so that $f:\mathbb{R}^3 \rightarrow \mathbb{R}^2$, with $f(u_0, \eta_0, t) = (u(t), \eta(t))$, and $\sigma_a = \sigma_r = -0.45$. The advantage of $M=1$ is that we have access to an analytic, cheap to compute solution for $f$. Of course, this particular case can also be solved using a classical ODE solver, which allows us to test it end to end. It can thus be generalized to higher dimensions ($M > 1$).

All neural network training instances have been performed in \texttt{Python}, with \texttt{Tensorflow}. We used a fully connected neural network with hyperparameters chosen using a simple grid search. The final values are: 2 hidden layers, relu activation function, and 32 units for each layer, trained with the Mean Squared Error (MSE) loss function using Adam optimization algorithm with a batch size of 50000, for 40000 epochs and on $N+N' = 50000$ points, with $N=N'$. We trained the model for $(u(t), \eta(t)) \in \mathbb{R}$, with the $N+N'$ points sampled uniformly (BS), and compared it to TBS applied on $N'$ after a uniform sampling of $N$ points (TBS). We did so for several values of $n$, $n_{\text{GMM}}$ and $\vepsilon=\epsilon(1,1,1)$, to fine tune these parameters. We finally select  $\epsilon = 5\times 10^{-4}$, $n=2$ and $n_{\text{GMM}} = 10$. The data points used in this case have been sampled with an explicit Euler scheme. Note that we used this scheme because it is a stable converging accurate scheme if the time steps for the resolution are fine enough (which we thoroughly checked). Depending on the application, other schemes could be used (faster ones, stabler ones etc.). As we here mainly aim at building a database of solution, we are not constrained by some computational restrictions. So we decided to use a very simple scheme, easy to handle which can easily produce accurate solutions, even if costly (as it is only an offline cost). This experiment has been repeated 50 times to ensure statistical significance of the results. 

\textbf{Table \ref{ode-table}} summarizes the MSE, i.e. the $L_2$ norm of the error of $f_{\vtheta}$ and $L_{\infty}$ norm, with $L_{\infty}(\vtheta) = \underset{\vx\in \mathbf{S}}{\max}(|f(\vx) - f_{\vtheta}(\vx)|)$ obtained at the end of the training phase. This last metric is important because the goal in computational physics is not only to be averagely accurate, which is measured with MSE, but to be accurate over the whole input space $\mathbf{S}$. Those norms are estimated using a same test data set of $N_{test} = 50000$ points. The values are the means of the $50$ independent experiments displayed with a $95\%$ confidence interval. These results reflect an error reduction of 6.6\% for $L_2$ and of 45.3\% for $L_{\infty}$, which means that TBS mostly improves the $L_{\infty}$ error of $f_{\vtheta}$. Moreover, the $L_{\infty}$ error confidence intervals do not intersect so the gain is statistically significant for this norm.

\begin{table}[h]
  \centering
  \begin{tabular}{lllll}
     Sampling & $L_2$ error $ (\times 10^{-4})$ & $L_{\infty}$ $ (\times 10^{-1})$ & AEG$ (\times 10^{-2})$ & AEL$ (\times 10^{-2})$\\
    \hline
    BS & $1.22 \pm 0.13 $ & $5.28 \pm 0.47$ & - & -    \\
    \textbf{TBS}     & $\boldsymbol{1.14} \pm 0.15 $ & $\boldsymbol{2.96} \pm 0.37 $  &  $2.51 \pm 0.07 $ & $0.42 \pm 0.008 $  \\
  \end{tabular}
  \caption{Comparison between \textcolor[rgb]{0.137,0.341,0.537}{BS} and \textcolor[rgb]{0.85,0.282,0}{TBS}. \label{ode-table}}
\end{table}

\begin{figure}[!h]
   \begin{subfigure}{0.5\textwidth}
     \includegraphics[trim={0cm 0cm -2cm 0.86cm},clip,width=1.0\linewidth]{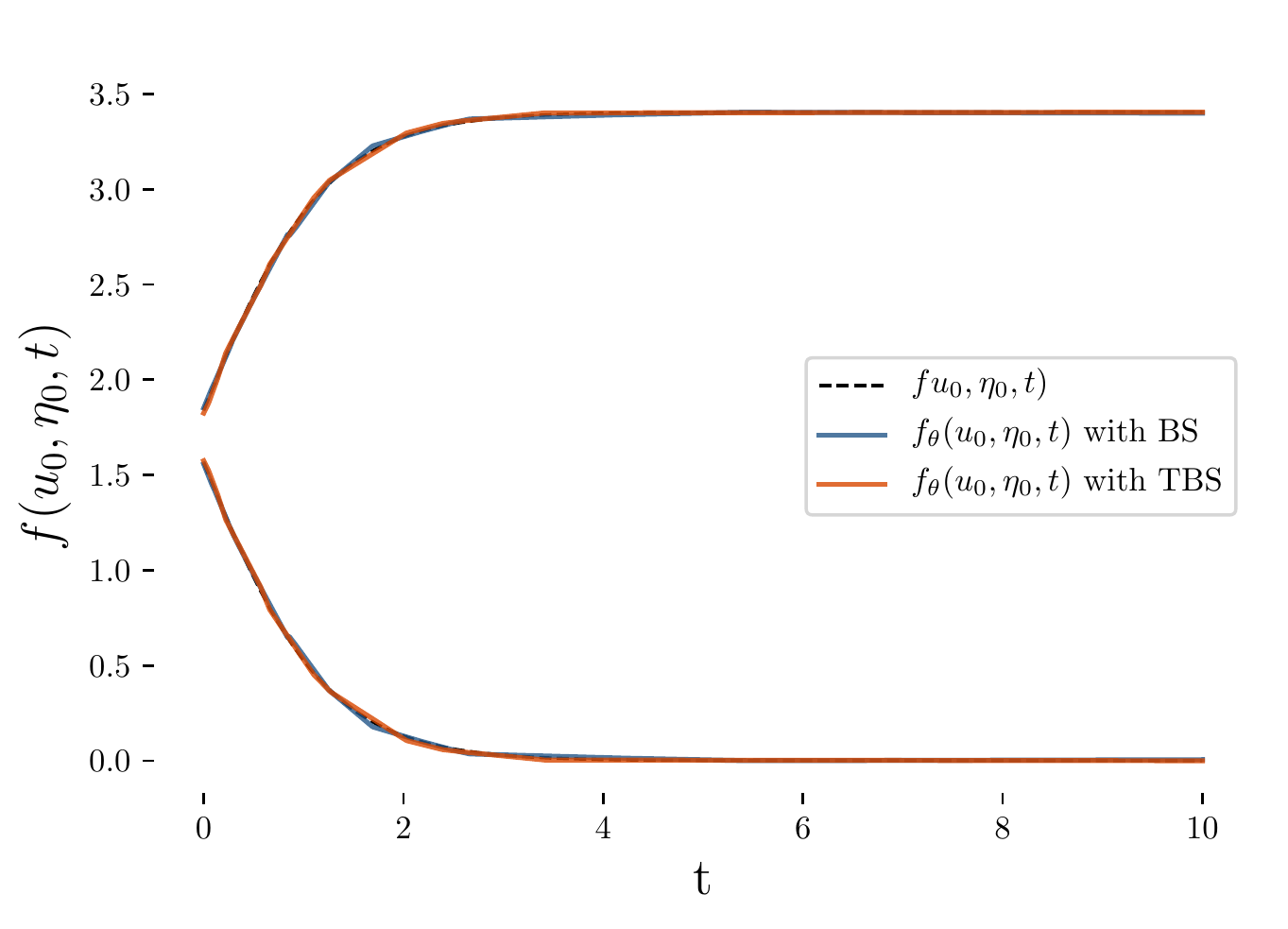}
     \caption{\label{fig:tbsa}}
     \vspace{0.4cm}
   \end{subfigure}
   \begin{subfigure}{0.5\textwidth}
     \includegraphics[trim={0cm 0cm -2cm 0.86cm},clip,width=1.0\linewidth]{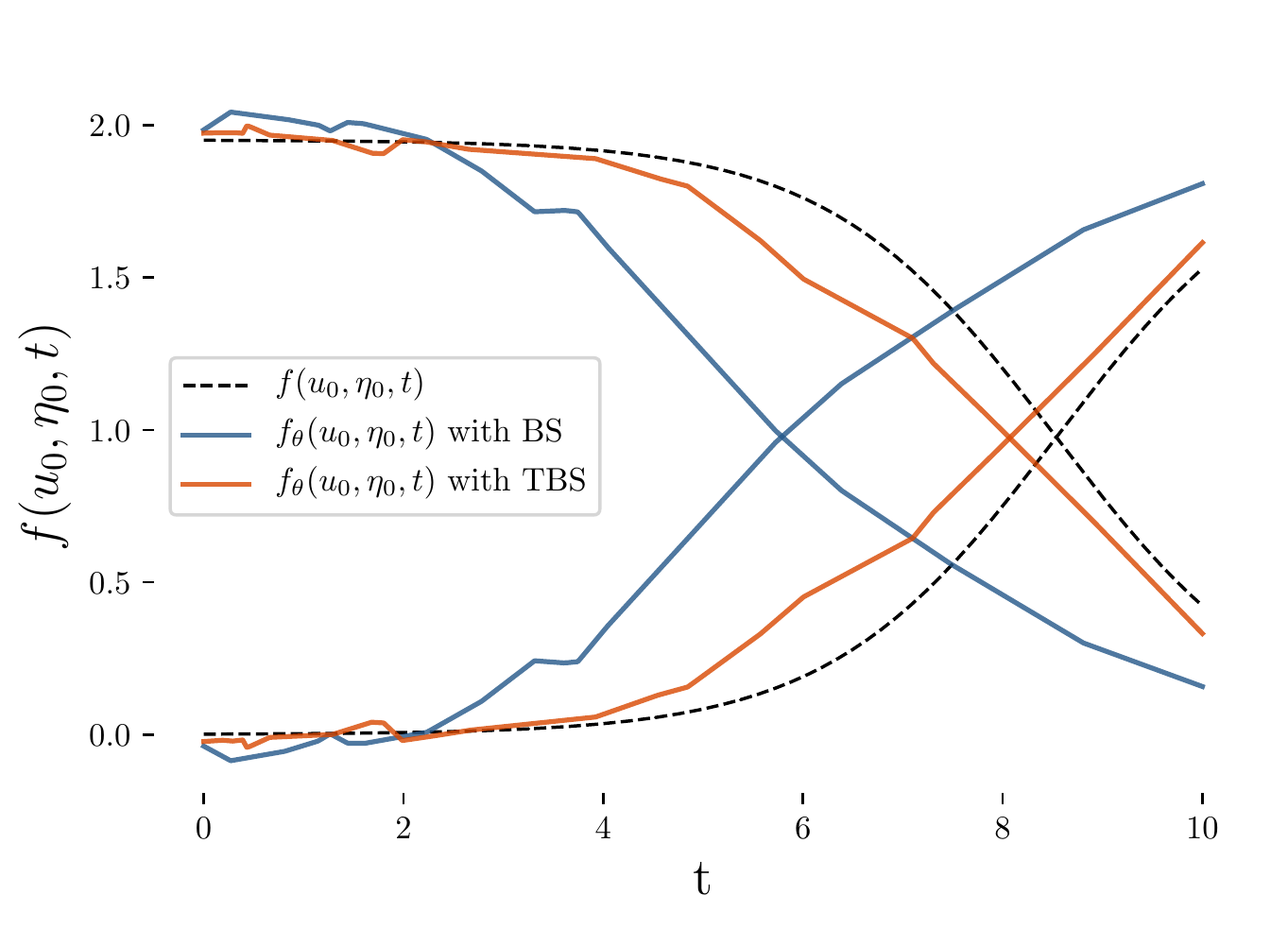}
     \caption{\label{fig:tbsb}}
     \vspace{0.4cm}
   \end{subfigure}
   \begin{subfigure}{0.5\textwidth}
     \centering
     \includegraphics[trim={0cm 0cm 0cm 0.86cm},clip,width=1.01\linewidth]{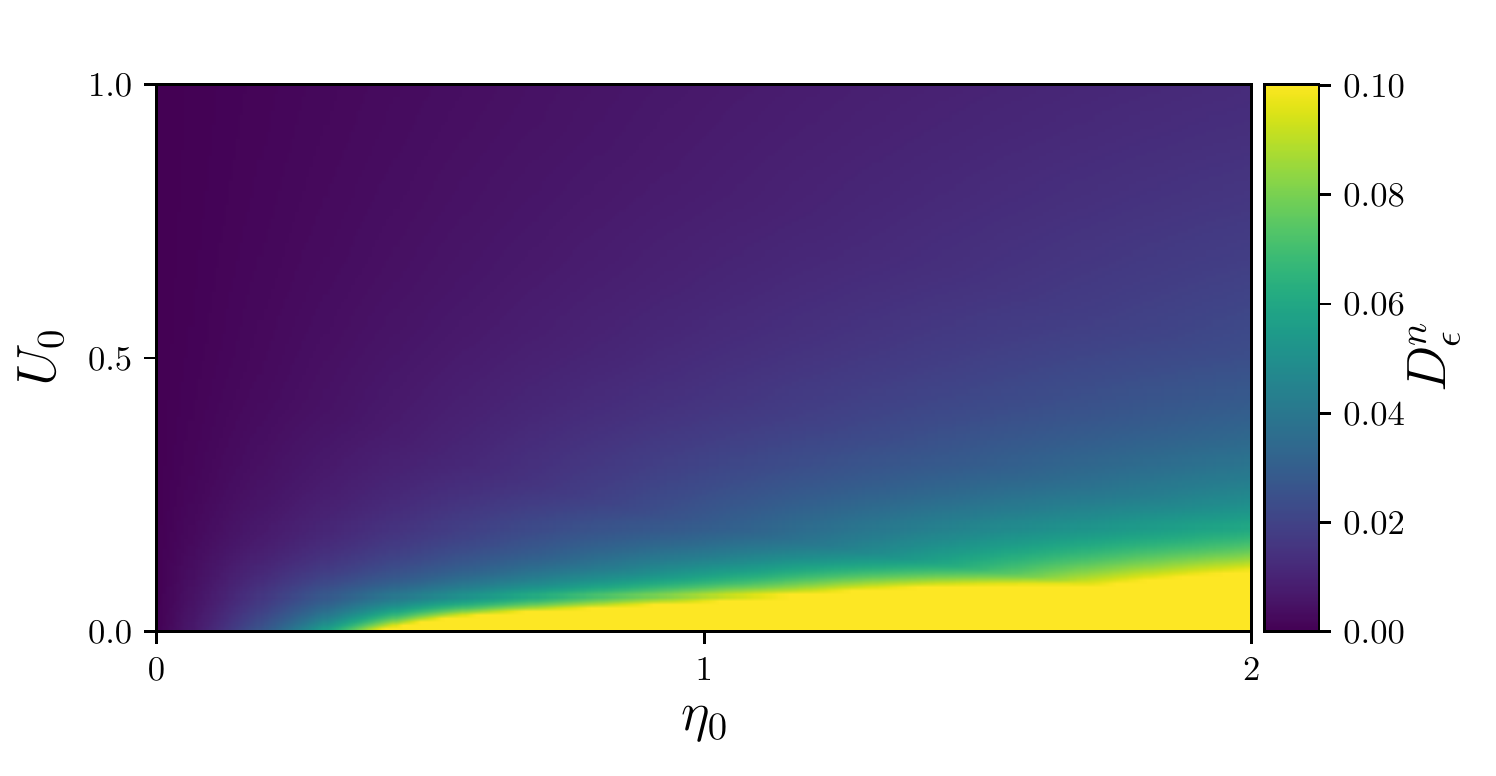}
     \caption{\label{fig:tbsc}}
   \end{subfigure}
  \begin{subfigure}{0.5\textwidth}
     \centering
     \includegraphics[trim={0cm 0cm 0cm 0.86cm},clip,width=1.01\linewidth]{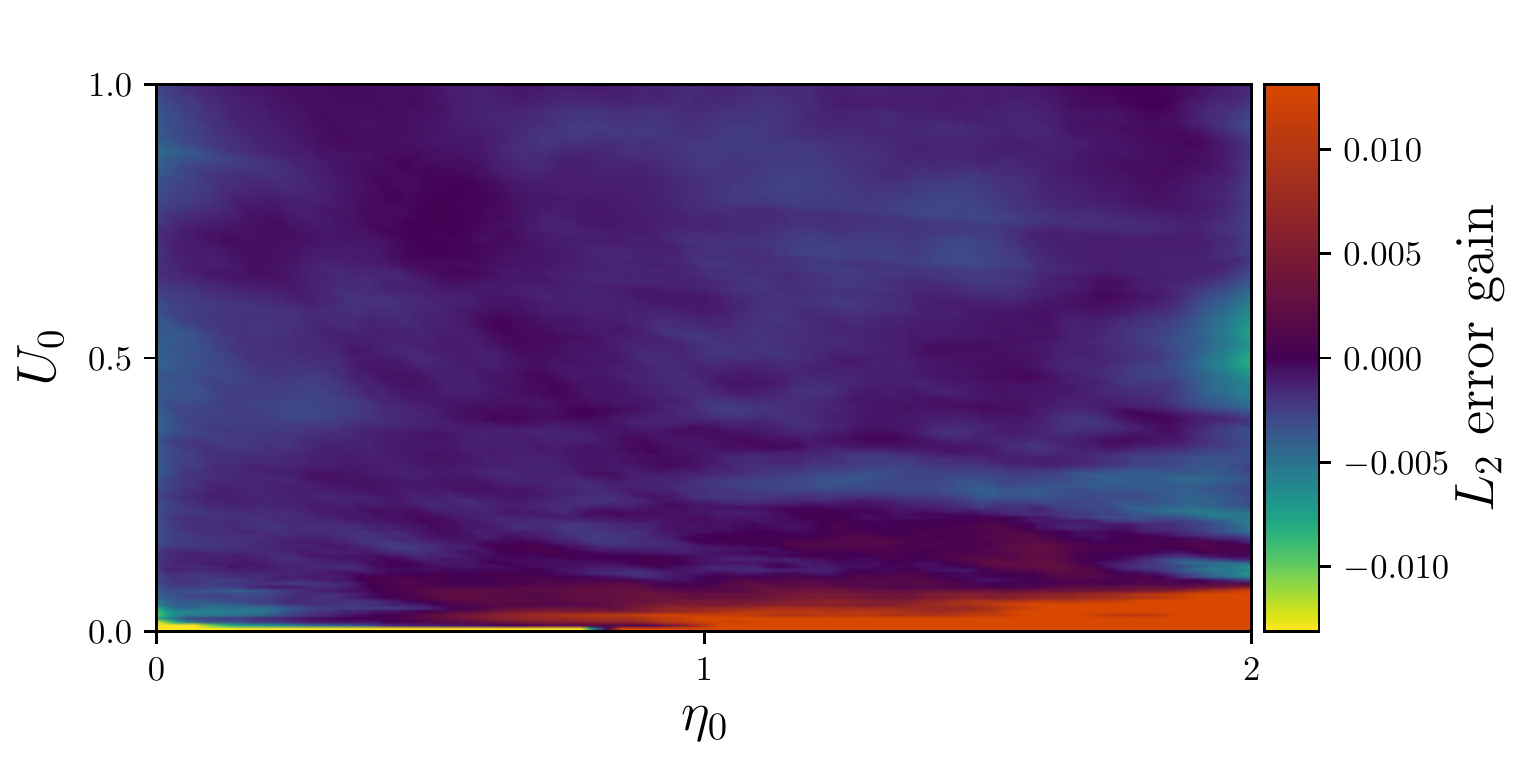}
     \caption{\label{fig:tbsd}}
   \end{subfigure}
   \caption{\textbf{(a) } $t \rightarrow f_{\vtheta}(u_0,\eta_0,t)$ for randomly chosen $(u_0,\eta_0)$, for $f_{\vtheta}$ obtained with the two samplings. \textbf{(b)} $t \rightarrow f_{\vtheta}(u_0,\eta_0,t)$ for $(u_0,\eta_0)$ resulting in the highest point-wise error with the two samplings. \textbf{(c)} $u_0,\eta_0 \rightarrow \underset{0 \leq t \leq 10} {\max} D^{n}_{\vepsilon}(u_0, \eta_0, t)$ w.r.t. $(u_0, \eta_0)$. \textbf{(d)} $u_0,\eta_0 \rightarrow g_{\theta_{BS}}(u_0,\eta_0) -g_{\theta_{TBS}}(u_0,\eta_0)$, }
\end{figure}

Figure \ref{fig:tbsa} shows how the neural network can perform for an average prediction. Figure \ref{fig:tbsb} illustrates the benefits of \textcolor[rgb]{0.85,0.282,0}{TBS} relative to \textcolor[rgb]{0.137,0.341,0.537}{BS} on the $L_{\infty}$ error (Figure 2b). These 2 figures confirm the previous observation about the gain in $L_{\infty}$ error.   
Finally, Figure \ref{fig:tbsc} displays $u_0,\eta_0 \rightarrow \underset{0 \leq t \leq 10} {\max} D^{n}_{\vepsilon}(u_0, \eta_0, t)$ w.r.t. $(u_0, \eta_0)$ and shows that $D^{n}_{\vepsilon}$ increases when $U_0 \rightarrow 0$. TBS hence focuses on this region. Note that for the readability of these plots, the values are capped to $0.10$. Otherwise only few points with high $D^{n}_{\vepsilon}$ are visible. Figure \ref{fig:tbsd} displays $u_0,\eta_0 \rightarrow g_{\theta_{BS}}(u_0,\eta_0) -g_{\theta_{TBS}}(u_0,\eta_0)$, with  $g_{\vtheta}:u_0,\eta_0 \rightarrow \underset{0 \leq t \leq 10} {\max}\|f(u_0,\eta_0,t) - f_{\vtheta}(u_0,\eta_0,t)\|_2^2$ where $\theta_{BS}$ and $\theta_{TBS}$ denote the parameters obtained after a training with BS and TBS, respectively.  It can be interpreted as the error reduction achieved with TBS. 

The highest error reduction occurs in the expected region. Indeed, more points are sampled where $D^{n}_{\vepsilon}$ is higher. The error is slightly increased in the rest of $\mathbf{S}$, which could be explained by a sparser sampling on this region. However, as summarized in Table \ref{ode-table}, the average error loss (AEL) of TBS is around six times lower than the average error gain (AEG), with $AEG = \mathbb{E}[Z(u_0,\eta_0)\mathbf{1}_{Z>0}]$ and $AEL = \mathbb{E}[Z(u_0,\eta_0)\mathbf{1}_{Z<0}]$ where $Z(u_0,\eta_0) =  g_{\theta_{BS}}(u_0,\eta_0) -g_{\theta_{TBS}}(u_0,\eta_0)$. In practice, AEG and AEL are estimated using uniform grid integration, and averaged on the $50$ experiments. 

\section{Generalization of Taylor based Sampling}

The previous section empirically validated the intuition behind the construction of a new, more efficient training distribution $d\mathbb{P}_{\bar{\rvx}}$. However, this new distribution cannot always be applied as-is for two reasons. \textbf{Problem 1:} $\{Df^{2}_{\vepsilon}(\vx_1),...,Df^{2}_{\vepsilon}(\vx_N)\}$ cannot be evaluated since it requires to compute the derivatives of $f$, and it assumes that $f$ is differentiable, which is often not true. Moreover, the previously described setting, in which we focus on $f$ derivatives, is not suited to classification tasks where the notion of derivatives is not straightforward. \textbf{Problem 2:} even if  $\{Df^{2}_{\vepsilon}(\vx_1),...,Df^{2}_{\vepsilon}(\vx_N)\}$ could be computed and new points sampled, we could not obtain their labels to complete the training data set. In this section, we alleviate this concern to be able to use insights from $d\mathbb{P}_{\bar{\rvx}}$ in practice.

\subsection{From Taylor expansion to local variance}
\label{sec:problem1}

To overcome \textbf{problem 1}, we construct a new metric based on statistical estimation. In this paragraph, $n_i >1$ but $n_o= 1$. The following derivations can be extended to $n_o > 1$ by applying it to $f$ element-wise and then taking the sum across the $n_o$ dimensions.

\begin{lemma}\label{estim}
Let $\rve \sim \mathcal{N}(0, \epsilon \mI_{n_i})$ with $\epsilon \in \mathbf{R}^+$ and $\mI_{n_i}$ the identity matrix of dimension $n_i$. Let $\vepsilon = \epsilon(1,...,1)$. Then,

\begin{equation*}
    Var(f(\vx + \rve) ) =  Df_{\vepsilon}^2(\vx) + \mathcal{O}(\|\vepsilon\|^3_2).
\end{equation*}
\end{lemma}

The demonstration can be found in \textbf{\hyperref[appA]{Appendix A}}. Using the unbiased estimator of variance, we thus define new indices $\widehat{Df_{\vepsilon}^2}(\vx)$ by

\begin{equation}\label{Dhat}
    \widehat{Df_{\vepsilon}^2}(\vx) = \frac{1}{k-1}\sum_{i = 1}^{k} \Big(f(\vx + \boldsymbol{\epsilon_i}) - f(\vx)\Big)^2,
\end{equation} 

with $\{\vepsilon_1,...,\vepsilon_k\}$ $k$ samples of $\vepsilon$. The metric $\widehat{Df^2_{\vepsilon}}(\vx) \underset{k \rightarrow \infty}{\rightarrow} Var(f(\vx + \vepsilon) )$ and  $Var(f(\vx + \vepsilon) ) = Df^2_{\vepsilon}(\vx) + \mathcal{O}(\|\vepsilon\|^3_2)$, so $\widehat{Df^2_{\vepsilon}}(\vx)$ is a biased estimator of $Df^2_{\vepsilon}(\vx)$, with bias $\mathcal{O}(\|\vepsilon\|^3_2)$. Hence, when $\vepsilon \rightarrow 0$, $\widehat{Df^2_{\vepsilon}}(\vx)$ becomes an unbiased estimator of $Df^2_{\vepsilon}(\vx)$. It is possible to compute $\widehat{Df^2_{\vepsilon}}(\vx)$ from any set of points centered around $\vx$. Therefore, we evaluate $\widehat{Df^2_{\vepsilon}}(\vx_i)$ for each $i \in \{1,...,N\}$ using the set $\mathcal{S}_k(\vx_i)$ of $k$-nearest neighbors of $\vx_i$. We note this metric $\widehat{Df^2}(\vx_i)$, where we replace $f(\vx_i + \boldsymbol{\epsilon_i})$ by $f(\vx_l)$, the values of $f$ for the neighbors of $\vx_i$ ($\vx_l\in \mathcal{S}_k(\vx_i)$) and $f(\vx_i)$ by $\frac{1}{k}\sum_{\vx_l \in \mathcal{S}_k(\vx_i)}^{k} f(\vx_l)$,  the average of $f$ on the neighbors of $\vx_i$ :\\

\begin{equation} \label{Dhat2}
    \widehat{Df^2}(\vx_i) = \frac{1}{k-1}\sum_{\vx_j \in \mathcal{S}_k(\vx_i)} \Big(f(\vx_j) - \frac{1}{k}\sum_{\vx_l \in \mathcal{S}_k(\vx_i)}^{k} f(\vx_l)\Big)^2,
\end{equation}

\Eqref{Dhat2} has several practical advantages. First, $\widehat{Df^2}$ can even be applied to non-differentiable functions and for classification problems, unlike \eqref{dn}. Second, the definition of $\widehat{Df^2}(\vx)$ does not rely on $\vepsilon$, unlike \eqref{Dhat}. To compute $\widehat{Df^2}$, all we need are $\{f(\vx_1),...,f(\vx_N)\}$, the points used for the training of the neural network. In addition, \eqref{Dhat2} can even be applied when the data points are too sparse for the nearest neighbors of $\vx_i$ to be considered as close to $\vx_i$, which is almost always the case in high dimension. It can thus be seen as a generalization of $\widehat{Df_{\vepsilon}^2}(\vx)$, which tends towards $Df_{\vepsilon}^2(\vx)$ locally.

\subsection{From sampling to weighting}
\label{sec:problem2}

To tackle \textbf{problem 2}, recall that the goal of the training is to find $\vtheta^* = \underset{\vtheta}{\operatorname{argmin}} \; \widehat{J_{\rvx}}(\vtheta)$, with $\widehat{J_{\rvx}}(\vtheta) = \frac{1}{N}\sum_{i} L(f(\vx_i),  f_{\vtheta}(\vx_i))$. With the new distribution based on previous derivations, the procedure is different. Since the training points are sampled using $\widehat{Df^2_{\vepsilon}}$, we no longer minimize $\widehat{J_{\rvx}}(\vtheta)$, but $\widehat{J_{\bar{\rvx}}}(\vtheta) = \frac{1}{N}\sum_{i} L(f(\bar{\vx}_i), f_{\vtheta}(\bar{\vx}_i))$, with $\bar{\rvx} \sim d\mathbb{P}_{\bar{\rvx}}$ the new distribution. However, $\widehat{J_{\bar{\rvx}}}(\vtheta)$ estimates

\begin{equation*}
    J_{\bar{\rvx}}(\vtheta) = \int_{\mathbf{S}} L(f(\vx), f_{\vtheta}(\vx)) d\mathbb{P}_{\bar{\rvx}}.
\end{equation*}

Let $p_{\rvx}(\vx)d\vx = d\mathbb{P}_{\rvx}$, $p_{\bar{\rvx}}(\vx)d\vx = d\mathbb{P}_{\bar{\rvx}}$ be the pdfs of $\rvx$ and $\bar{\rvx}$ (note that $Df^2_{\vepsilon} \propto p_{\bar{\rvx}}$). Then,

\begin{equation*}
    J_{\bar{\rvx}}(\vtheta) =  \int_{\mathbf{S}} L(f(\vx), f_{\vtheta}(\vx)) \frac{p_{\bar{\rvx}}(\vx)}{p_{\rvx}(\vx)}d\mathbb{P}_{\rvx}.
\end{equation*}

The straightforward Monte Carlo estimator for this expression of $J_{\bar{\rvx}}(\vtheta) $ is 

\begin{equation}\label{J2*}
\begin{split}
    \widehat{J_{\bar{\rvx}}}(\vtheta) &= \frac{1}{N} \sum_{i} L(f(\vx_i), f_{\vtheta}(\vx_i))\frac{p_{\bar{\rvx}}(\vx_i)}{p_{\rvx}(\vx_i)}
    \propto \frac{1}{N} \sum_{i} L(f(\vx_i), f_{\vtheta}(\vx_i))\frac{\widehat{Df^2}(\vx_i)}{p_{\rvx}(\vx_i)}.
\end{split}
\end{equation}

Thus, $J_{\bar{\rvx}}(\vtheta) $ can be estimated with the same points as $J_{\rvx}(\vtheta)$ by weighting them with $w_i = \frac{\widehat{Df^2}(\vx_i)}{p_{\rvx}(\vx_i)}$. 

The expression of $w_i$ involves $p_{\rvx}$, the distribution of the data. Just like for $f$, we do not have access to $p_{\rvx}$. The estimation of $p_{\rvx}$ is a challenging task by itself, and standard density estimation techniques such as \textit{K}-nearest neighbors or Gaussian Mixture density estimation led to extreme estimated values of $p_{\rvx}(\vx_i)$ in our experiments. Therefore, we decided to only apply $\omega_i = \widehat{Df^2}(\vx_i)$ as a first-order approximation. In practice, we re-scale the weights between $1$ and $m$, a hyperparameter, and then divide them by their sum to avoid affecting the learning rate. 

As a result, we obtain a new methodology based on weighting the training data set. We call this methodology Variance Based Sample Weighting (VBSW).

\section{Variance Based Sample Weighting}

In this part, we sum up Variance Based Sample Weighting (VBSW) to clarify its application to machine learning problems. We also study this methodology through toy experiments.

\subsection{Methodology}

Variance Based Samples Weighting (VBSW) is recapitulated in Algorithm \ref{alg:vbsw}. \textbf{Line 1:} $m$ and $k$ are hyperparameters that can be chosen jointly with all other hyperparameters, e.g. using a random search. Their effects and interactions are studied and discussed in Sections \ref{toy} and \ref{robustness}. \textbf{Line 2-3:} \eqref{Dhat2} is applied to compute the weights $w_i$ that are used to weight the data set. Notations $\{(w_1, \vx_1), ... , (w_N,\vx_N)\}$ denote that each $\vx_i$ is weighted by $w_i$. To perform a nearest-neighbors search, we use an approximate nearest neighbor search technique called hierarchical navigable small world graphs \citep{hnsw} implemented by nmslib \citep{nmslib}. \textbf{Line 4:} Train $f_{\vtheta}$ on the weighted data set.

\begin{algorithm}[h!]
    \caption{ \small Variance Based Samples Weighting (VBSW)}
    \label{alg:vbsw}
     {\bfseries Inputs: } $k$, $m$\;
     Compute $\{\widehat{Df^{2}}(\vx_1),...,\widehat{Df^{2}}(\vx_N)\}$ using \eqref{Dhat2}\;
     Construct a new training data set $\{(w_1, \vx_1), ... , (w_N,\vx_N)\}$\;
     Train $f_{\vtheta}$ on $\{(w_1, f(\vx_1)), ... , (w_N, f(\vx_N))\}$ \;
  
  \end{algorithm}

\subsection{Toy experiments \& hyperparameter study }\label{toy}
VBSW is studied on a Double Moon (DM) classification problem, the Boston Housing (BH) regression, and Breast Cancer (BC) classification data sets.


\begin{figure}[h!]
    \centering
    \begin{subfigure}{0.3\textwidth}
        \includegraphics[trim=2cm 0.5cm 1cm 0.5cm,width=1.0\linewidth]{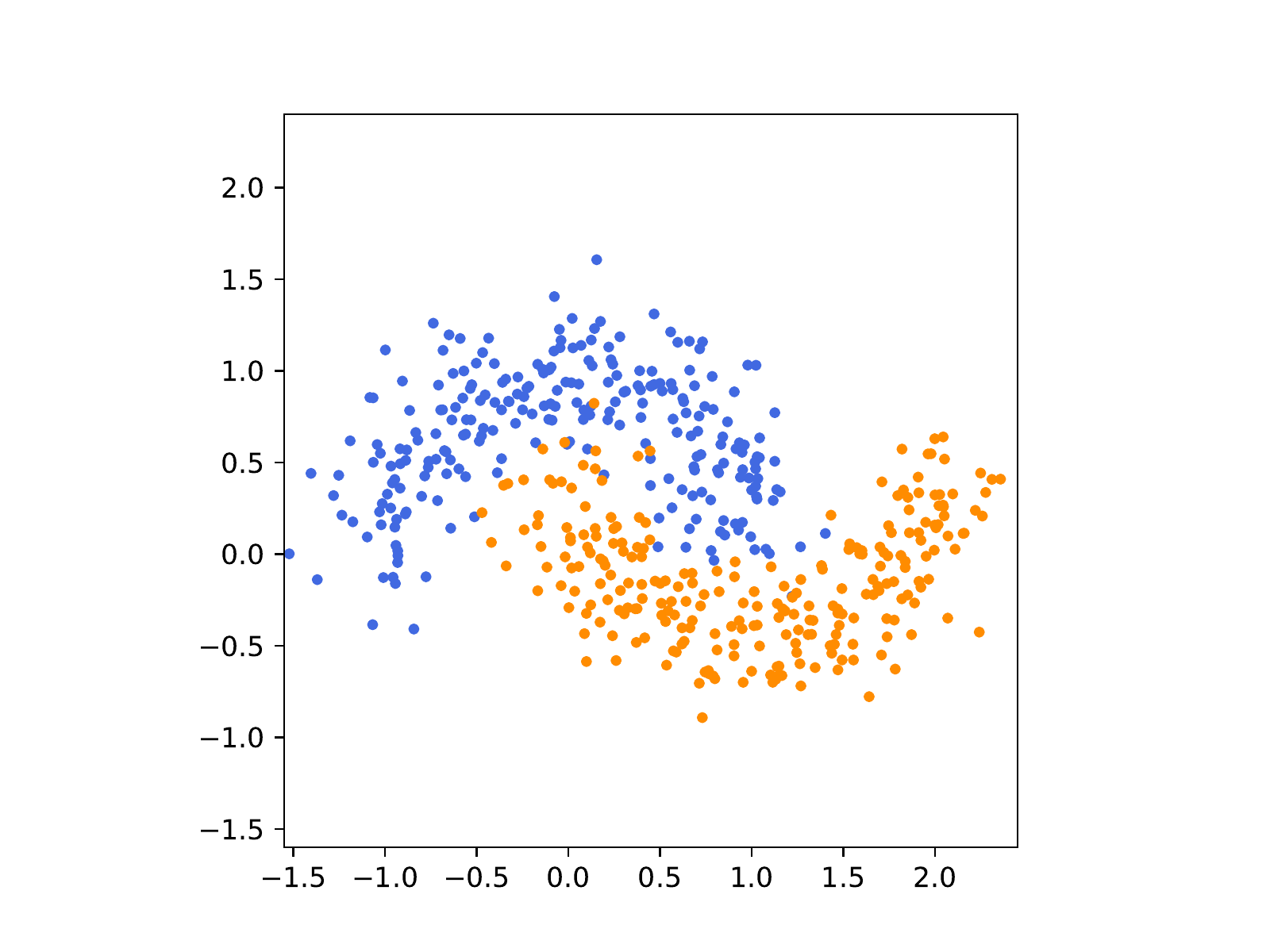}
        \caption{\label{doublema}}
    \end{subfigure}   
    \begin{subfigure}{0.3\textwidth}
        \includegraphics[trim=2cm 0.5cm 1cm 0.5cm,width=1.0\linewidth]{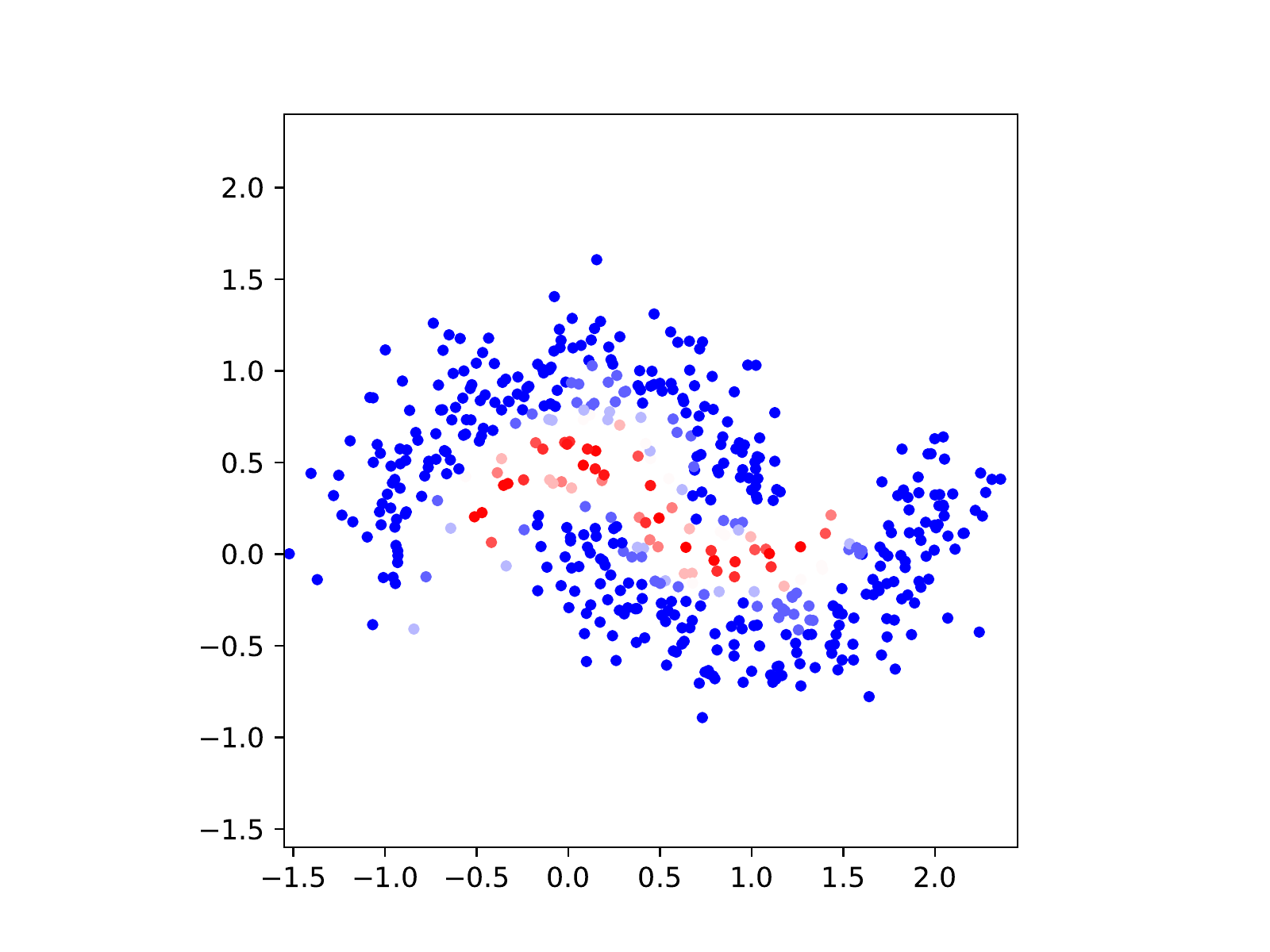}
        \caption{\label{doublemb}}
    \end{subfigure}

\caption{ \small From left to right: \textbf{(a)} Double Moon (DM) data set.  \textbf{(b)} Heat map of the value of $w_i$ for each $\vx_i$ (red is high and blue is low) }
\label{doublem}

\end{figure}

For DM, Figure \ref{doublemb} shows that the points with higher $w_i$ (in red) are close to the boundary between the two classes. Indeed, in classification, VBSW can be interpreted as a local label agreement. This behavior verifies recent findings of \cite{understandingIW} where authors conclude that in classification, a good set of weights would put importance on points close to the decision boundary. 

We train a Multi-Layer Perceptron of $1$ layer of $4$ units, using Stochastic Gradient Descent (SGD) and binary cross-entropy loss function, on a $300$ points training data set for $50$ random seeds. In this experiment, VBSW, i.e. weighting the data set with $w_i$ is compared to the baseline where no weights are applied. The results of Table \ref{test-table} show the improvement obtained with VBSW. 

\begin{table}[h!]

    \centering
        \begin{tabular*}{0.5\textwidth}{lll}
            & VBSW & baseline  \\
            \hline
            DM     & $\textbf{99.4}$, $\textbf{94.44} \pm 0.78$ & $99$, $92.06 \pm 0.66 $  \\
            BH & $\textbf{13.31}$, $\textbf{13.38} \pm 0.01$  & $14.05$, $14.06 \pm 0.01$    \\
            BC & $\textbf{99.12}$, $97.6 \pm 0.34$  & $98.25$, $97.5 \pm 0.11$    \\
            
        \end{tabular*}
        \caption{\small \label{test-table} \textbf{best, mean + se} for each method. The metric used is accuracy for DM and BC and Mean Squared Error for BH.}
    
    \end{table}
For BH data set, a linear model is trained, and for BC data set, an MLP of $1$ layer and $30$ units, with a train-validation split of $80\% - 20\%$. Both models are trained with Adam \cite{adam}. Since these data sets are small and the models are light, we study the effects of $m$ and $k$ on the error. Moreover, BH is a regression task and BC a classification task, so it allows studying the effect of hyperparameters more extensively. 

For BH and BC experiments, we conduct a grid search for VBSW on the values of $m$ and $k$. As a reminder, $m$ is the ratio between the highest and the lowest weights, and $k$ is the number of neighbor points used to compute the local variance. We train a linear model for BH and a MLP with $30$ units for BC with VBSW on a grid of $20$ values of $m$ equally distributed between $2$ and $100$ and $20$ values of $k$ equally distributed between $10$ and $50$. As a result, we train the model on $400$ pairs of $(m, k)$ values and with $10$ different random seeds for each pair. 

\begin{figure}[!h]
   \begin{subfigure}{0.49\textwidth}
     \includegraphics[width=1.0\linewidth]{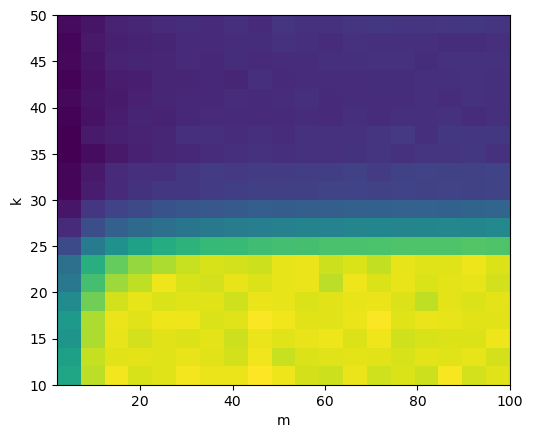}
     
   \end{subfigure}
   \begin{subfigure}{0.49\textwidth}
     \includegraphics[width=1.0\linewidth]{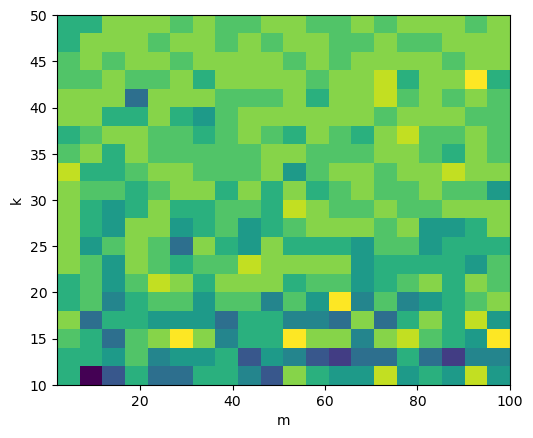}
     
   \end{subfigure}
   \caption{ Color map of the error, with respect to $m$ and $k$. \textbf{Left:} BH data set, for the mean of the MSE across $10$ different seeds and \textbf{right:} BC data set, for the mean of $1 - acc$ across these seeds. Blue is lower.}
   \label{hyp-search}
\end{figure}

These experiments, illustrated in Figure \ref{hyp-search} show that the influence of $m$ and $k$ on the performances of the model can be different. For BH data set, low values of $k$ clearly lead to poorer performances. Hyperparameter $m$ seems to have less impact, although it should be chosen not too far from its lowest value, $2$.  For BC data set, on the contrary, the best performances are obtained for low values of $k$, while a high value could be chosen for $m$. These experiments highlight that the impact of $m$ and $k$ can be different between classification and regression, but it could also be different depending on the data set. Hence, we recommend considering these hyperparameters like many others involved in deep learning, selecting their values using hyperparameters optimization techniques. 

It also shows that many different $(m,k)$ pairs lead to error improvement. It suggests that the weights approximation does not have to be exact for VBSW to be effective, as stated in Section \ref{robustness}.

\subsection{Cost efficiency of VBSW}

VBSW's computational burden mostly relies on the complexity of the nearest neighbor search algorithm, which is independent and can be used as a third-party algorithm. When the data set is not too large, classical techniques like KDtree \citep{kdtree} can be used. However, when the number of points and the dimension of the data set increase, approximate nearest neighbors searches may be necessary to keep satisfying performances. In the previous examples, KDtree is more than sufficient. However, since we deal with more complex examples in the following, we directly use nmslib \citep{nmslib}, an approximate nearest neighbors search library for homogeneity of the implementation.

\section{VBSW for deep learning}
\label{sec:exp_vbsw}

 The high dimensionality of many deep learning problems makes VBSW difficult to apply in the form previously described. In this part, we adapt VBSW to such problems and study its application to various real-world learning tasks. We also study the robustness of VBSW and its complementarity with other similar techniques.

\subsection{Methodology}

We mentioned that local variance could be computed using already existing points. This statement implies finding the nearest neighbors of each point. In extremely high-dimensional spaces like image spaces, the curse of dimensionality makes nearest neighbors vacuous. In addition, the data structure may be highly irregular, and the concept of nearest neighbor may be misleading. Thus, it would be irrelevant to evaluate $\widehat{Df^2}$ directly on this data.

One of the strengths of deep learning is to construct good representations of the data embedded in lower-dimensional latent spaces. For instance, in Computer Vision, convolutional neural networks' deeper layers represent more abstract features. We could leverage this representational power of neural networks and simply apply our methodology within this latent feature space.

Variance Based Samples Weighting (VBSW) for deep learning is recapitulated in Algorithm \ref{alg:vbswdl}. Here,  $\mathcal{M}$ is the initial neural network whose feature space will be used to project the training data set and apply VBSW. \textbf{Line 1:} $m$ and $k$ are hyperparameters that can be chosen jointly with all other hyperparameters, e.g. using a random search. Their effects and interactions are studied and discussed in Sections \ref{toy} and \ref{robustness}. \textbf{Line 2:} The initial neural network, $\mathcal{M}$, is trained as usual. Notations $\{(\frac{1}{N},\vx_1), ... , (\frac{1}{N}, \vx_N)\}$ is equivalent to $\{\vx_1, ... , \vx_N\}$, because all the weights are the same ($\frac{1}{N}$). \textbf{Line 3:} The last fully connected layer is discarded, resulting in a new model $\mathcal{M^*}$, and the training data set is projected in the feature space. \textbf{Line 4-5:} \eqref{Dhat2} is applied to compute the weights $w_i$ that are used to weight the projected data set. \textbf{Line 6:} The last layer is re-trained (which is often equivalent to fitting a linear model) using the weighted data set and added to $\mathcal{M^*}$ to obtain the final model $\mathcal{M}_f$. As a result, $\mathcal{M}_f$ is a composition of the already trained model $\mathcal{M^*}$ and $f_{\vtheta}$ trained using the weighted data set.

\begin{algorithm}[h!]
    \caption{ \small Variance Based Samples Weighting (VBSW) for deep learning}
    \label{alg:vbswdl}
     {\bfseries Inputs: } $k$, $m$, $\mathcal{M}$\;
     Train $\mathcal{M}$ on the training set $\{(\frac{1}{N},\vx_1), ... , (\frac{1}{N}, \vx_N)\}$\; $\{(\frac{1}{N},f(\vx_1)), ... , (\frac{1}{N},f(\vx_N))\}$\;
     Construct $\mathcal{M^*}$ by removing its last layer \;
     Compute $\{w_1 = \widehat{Df^{2}}(\mathcal{M^*}(\vx_1)),...,w_N = \widehat{Df^{2}}(\mathcal{M^*}(\vx_N))\}$ using \eqref{Dhat2}\;
     Construct a new training data set $\{(w_1, \mathcal{M^*}(\vx_1)), ... , (w_N,\mathcal{M^*}(\vx_N))\}$\;
     Train $f_{\vtheta}$ on the training set of inputs $\{(w_1, \mathcal{M^*}(\vx_1)), ... , (w_N,\mathcal{M^*}(\vx_N))\}$ with outputs $\{f(\vx_1), ... , f(\vx_N)\}$ and add it to $\mathcal{M^*}$. The final model is $\mathcal{M}_f$ = $f_{\vtheta}\circ  \mathcal{M^*}$\;
  
  \end{algorithm}

\subsection{Image Classification}\label{cifar10}

In this section, we study the performances of VBSW on MNIST \cite{mnist} and Cifar10 \cite{cifar} image classification data sets. For MNIST, we train LeNet \citep{lenet}, with $40$ different random seeds, and then apply VBSW for $10$ different random seeds, with Adam optimizer and categorical cross-entropy loss. Note that in the following, Adam is used with the default parameters of its \texttt{keras} implementation. We record the best value obtained from the $10$ VBSW training. We follow the same procedure for Cifar10, except that we train a ResNet20 for $50$ random seeds and with data augmentation and learning rate decay. The networks have been trained on 4 Nvidia K80 GPUs. The values of the hyperparameters used can be found in \textbf{\hyperref[appB]{Appendix B}}. We compare the test accuracy between LeNet 5 + VBSW, ResNet20 + VBSW, and the initial test accuracies of LeNet 5 and ResNet20 (baseline) for each of the initial networks. 

\begin{table}[h!]

\centering
    \begin{tabular*}{\textwidth}{l@{\extracolsep{\fill}}llll}
        & VBSW & baseline & gain per model \\
        \hline
        MNIST     & $\textbf{99.09}$, $\textbf{98.87} \pm 0.01$ & $98.99$, $98.84 \pm 0.01$  & $\textbf{0.15}$, $\textbf{0.03} \pm 0.01$  \\
        Cifar10 & $\textbf{91.30}$, $\textbf{90.64} \pm 0.07$  & $91.01$, $90.46 \pm 0.10$  & $\textbf{1.65}$, $\textbf{0.15} \pm 0.04$  \\
        \\
    \end{tabular*}
    
    \caption{\small \label{tb1} \textbf{best, mean + se} for each method. The metric used is accuracy.  For a model $\mathcal{M}$, the gain $g$ for this model is given by $g = \underset{1\leq i \leq10}{\operatorname{max}} (acc(\mathcal{M}^i_f) - acc(\mathcal{M}))$ with $acc$ the accuracy and $\mathcal{M}^i_f$ the VBSW model trained at the $i$-th random seed.}
    
\end{table}

\begin{figure}[h!]
    \centering
    \includegraphics[width=1.0\linewidth]{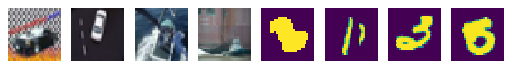}
    
    \caption{\small Samples from Cifar10 and MNIST with high $w_i$. Those pictures are either unusual or difficult to classify, even for a human (especially for MNIST).}
    \label{difficult}
\end{figure}

The results statistics are gathered in Table \ref{tb1}, which also displays statistics about the gain due to VBSW for each model. The results on MNIST are slightly but consistently better than for the baseline, by $0.1\%$ for the best with up to $0.15\%$ of accuracy gain per model. For Cifar10, we get a $0.3\%$ accuracy improvement for the best model and up to $1.65\%$ accuracy gain, meaning that among the $50$ ResNet20s, there is one whose accuracy has been improved by $1.65\%$ using VBSW. Note that applying VBSW took less than 15 minutes on a laptop with an i7-7700HQ CPU. A visualization of the samples weighted by the highest $w_i$ is given in Figure \ref{difficult}. 

\subsection{Text Classification and Regression}

In this section, we study the performances of VBSW on RTE and MRPC, two text classification data sets, and STS-B, a text classification data set, extracted from the glue benchmark \cite{glue}. For this application, we use Bert, a modern neural network based on transformers \cite{transformers} that is the state-of-the-art of text-based machine learning tasks. We do not pre-train Bert, like in the previous experiments, since it has been originally built for Transfer Learning purposes. Therefore, its purpose is to be used as-is and then fine-tuned on any text data set see \cite{bert}. However, because of the small size of the data set and the high number of model parameters, we chose not to fine-tune the Bert model and only to use the representations of the data sets in its feature space to apply VBSW. More specifically, we use tiny-bert \cite{tinybert}, which is a lighter version of the initial Bert. We train the linear model with TensorFlow to be able to add the trained model on top of the Bert model and obtain a unified model. RTE and MRPC are classification tasks, so we use binary cross-entropy loss function to train our models. STS-B is a regression task, so the model is trained with Mean Squared Error. All the models are trained with Adam optimizer. For each task, we compare the training of the linear model with VBSW and without VBSW (baseline). The results obtained with VBSW are better overall, except for Pearson Correlation in STS-B, which is slightly worse than baseline (Table \ref{tb2}).

\begin{table}[h!]
  \centering
  \resizebox{\textwidth}{!}{
    \begin{tabular}{llllll@{\extracolsep{\fill}}}
        & \multicolumn{2}{c}{VBSW} & \multicolumn{2}{c}{baseline}  \\
        \hline 
        & m1 & m2 & m1 & m2  \\
        \hline
        RTE & $\textbf{61.73}$, $\textbf{58.46} \pm 0.15$   & - & $61.01$, $58.09 \pm 0.13$ & -  \\ 
        STS-B  & $\textbf{62.31}$, $\textbf{62.20} \pm 0.01$   & $\textbf{60.99}$, $60.88 \pm 0.01$  & $61.88$, $61.87 \pm 0.01$ & $60.98$, $\textbf{60.92} \pm 0.01$  \\
        MRPC & $\textbf{72.30}$, $\textbf{71.71} \pm 0.03$   & $\textbf{82.64}$, $\textbf{80.72} \pm 0.05$ & $71.56$, $70.92 \pm 0.03$ & $81.41$, $80.02 \pm 0.07$ \\ 
        \\
    \end{tabular}}
    \caption{\small \label{tb2} \textbf{best, mean + se} for each method. For RTE the metric used is accuracy (m1). For STS-B, metric 1 (m1) is Spearman correlation and metric 2 (m2) is Pearson correlation. For MRPC, metric 1 (m1) is accuracy and metric 2 (m2) is F1 score.}
\end{table}

\subsection{Robustness of VBSW}\label{robustness}

In this section, we assess the robustness of VBSW. First, we focus on the robustness to label noise. To that end, we train a ResNet20 on Cifar10 with four different noise levels. We randomly change the label of $p \%$ training points for four different values of $p$ ($10$, $20$, $30$ and $40$). We then apply VBSW $30$ times and evaluate the obtained neural networks on a clean test set. The results are gathered in Table \ref{tab:noise}.

\begin{table}[h]
\resizebox{\textwidth}{!}{%
  \centering
  \begin{tabular}{lllll}
    noise  & $10\%$ & $20\%$ & $30\%$ & $40\%$ \\
    \hline
    original error & $87.43$ & $85.75$ & $84.05$ & $81.79$ \\
    VBSW & $\textbf{87.76}$,  $8\textbf{7.63}\pm 0.01 $ & $\textbf{86.03}$,  $\textbf{85.89}\pm 0.01 $ & $\textbf{84.35}$,  $\textbf{84.18}\pm 0.02 $ & $\textbf{82.48}$,  $\textbf{82.32}\pm 0.02 $  \\
  \end{tabular}}
  \caption{\label{tab:noise} \textbf{best, mean + se} of the training of a ResNet20 on Cifar10 for different label noise levels. These results illustrate the robustness of VBSW to labels noise.}
\end{table}

The results show that VBSW is still effective despite label noise. This specificity must be related to the robustness of VBSW with respect to the choice of hyperparameters $m$ and $k$, as seen in Section \ref{toy}. Indeed, it shows that many combinations of $m$ and $k$ improves the performances of the neural network, and therefore that VBSW is actually robust to error in the weights evaluation. Its robustness to label noise hence stems from its robustness to weights evaluation error, since label noise essentially hurts the accuracy of the weights evaluation.

Although VBSW is robust to label noise, note that the goal of VBSW is not to address noisy label problem, like discussed in Section \ref{sec:related_vbsw}. It may be more effective to use a sampling technique tailored specifically for this situation.

\subsection{Complementarity of VBSW}\label{complementarity}

Existing techniques based on dataset processing can be used jointly with VBSW, by applying the first technique during the initial training of the neural network and then applying VBSW on its feature space. To illustrate this specificity, we compare VBSW with the recently introduced Active Bias (AB) \citep{Chang} and transfer-learning-based curriculum learning (TCL) \citep{transfer-curriculum-learning}. AB dynamically weights the samples based on the variance of the probability of prediction of each point throughout the training, and TCL creates a curriculum based on sample difficulty evaluated on previously trained neural networks. Here, we study the effects of AB and TCL combined with VBSW for the training of a ResNet20 on Cifar10. Table \ref{tab:ab} gathers the results of experiments for different baselines: vanilla, for regular training with Adam optimizer, AB / TCL for training with AB / TCL, VBSW for the application of VBSW on top of regular training, and VBSW + AB / VBSW + CL for initial training with AB / TCL and the application of VBSW. Unlike in Section \ref{cifar10}, we do not use data augmentation nor learning rate decay in order to simplify the experiments.
\setlength\tabcolsep{3pt}
\begin{table}[h]
  \centering
  \begin{tabular}{lll}
     & accuracy ($\%$) & VBSW gpm\\
    \hline
    vanilla & $75.88$, $74.55\pm 0.11 $ & -\\
    AB & $76.33$, $75.14\pm 0.09 $  &- \\
    TCL & $78.54$, $77.46\pm 0.07 $  &- \\
    VBSW & $76.57$, $74.94\pm 0.10 $  & $0.94$, $0.40 \pm 0.03  $ \\
    AB + VBSW & $76.60$, $75.33 \pm 0.09 $ & $0.40$, $0.14 \pm 0.02  $\\
    TCL + VBSW & $\boldsymbol{79.86}$, $\boldsymbol{78.71} \pm 0.09 $ & $\boldsymbol{2.19}$, $\boldsymbol{1.26} \pm 0.08  $
  \end{tabular}
  \caption{\label{tab:ab} \textbf{Best, mean + se} of the training of 60 ResNet20s on Cifar10 for vanilla, VBSW, AB and AB + VBSW. The gain per model (gpm)  $g$ is defined by $g = \underset{1\leq i \leq10}{\operatorname{max}} (acc(\mathcal{M}^i_f) - acc(\mathcal{M}))$ with $acc$ the accuracy and $\mathcal{M}^i_f$ the VBSW model trained at the $i$-th random seed.}
\end{table}

The accuracy obtained with VBSW is quite similar to AB. While TCL yields better results than VBSW alone, the best accuracy is obtained when they are used jointly. Overall, the best neural networks are obtained when AB and TCL are used along with VBSW (AB + VBSW and TCL + VBSW), which demonstrates the complementarity of VBSW with other dataset processing techniques. Note that VBSW works much better when applied to a neural network initially trained with TCL. It means that TCL creates neural network features particularly suited to VBSW. This lead might be explored in future works.

\section{Discussion and Perspectives}

By studying the training distribution of the neural network, we explored a practical and classical question that naturally arises when performing surrogate modeling for approximating computer codes: how to construct the training set? We found that exploring this question led to findings that are also relevant for approximation theory, which is an important component of machine learning. 

Hence, the results obtained in this paper are impactful both for machine learning in numerical simulations and machine learning in general.

\subsection{Impact for numerical simulations}

 This work comes from the observation that, on our approximation problems, neural networks are more efficient when more data are sampled where the function to learn is steeper. It is an attempt to formalize this observation and to construct a workable methodology out of it. As a result, the methodologies for constructing the distribution $d\mathbb{P}_{\bar{\rvx}}$ can be used as new, principled designs of experiments.

In the context of numerical simulations, once $d\mathbb{P}_{\bar{\rvx}}$ is constructed, it is possible to sample new data from it. It alleviates \textbf{Problem 2}, described in Section \ref{sec:problem2}. In theory, \textbf{Problem 1} is also solved since we could have access to the derivatives - either by instrumenting the code with automatic differentiation if we have access to its implementation or by estimating them with finite differences. However, the implementation of automatic differentiation can be tedious, and if the computer code is slow and high dimensional, finite differences may be unaffordable. In that case, it is possible to use a third methodology based on the approximation of $\{Df^{2}_{\vepsilon}(\vx_1),...,Df^{2}_{\vepsilon}(\vx_N)\}$ using local variance, like VBSW, and the sampling of new points, like TBS. 

Finally, the method allows improving the error of neural networks without increasing the computational cost of their prediction. This achievement is of interest when they are intended to accelerate numerical simulations.

\subsection{Impact for machine learning}

 VBSW is validated on several tasks, complementary with other training distribution modification frameworks, and robust to noise. It makes it quite versatile. Moreover, the problem of high dimensionality and irregularity of $f$, which often arises in deep learning problems, is alleviated by focusing on the latent space of neural networks. This makes VBSW scalable. As a result, VBSW can be applied to complex neural networks such as ResNet, or Bert, for various machine learning tasks. 

The experiments support an original view of the learning problem that involves the local variations of $f$. The studies of Section \ref{sec:tbs}, that use the derivatives of the function to learn to sample a more efficient training data set, support this approach as well. This view is also bolstered up by conclusions of \cite{understandingIW}. VBSW allows extending this original view to problems where the derivatives of $f$ are not accessible and sometimes not defined. Indeed, VBSW comes from Taylor expansion, which is specific to differentiable functions, but in the end, it can be applied regardless of the properties of $f$. 

Finally, this method is cost-effective. In most cases, it allows to quickly improve the performances of a neural network using a regular CPU. It is better than carrying on entirely new training with a wider and deeper neural network.

\subsection{Further studies}

Although VBSW uses theoretically justified approximations concerning TBS, the actual effect of these approximations should be more thoroughly investigated. For instance, we could further study the impact of not explicitly using $p_{\rvx}$, the data distribution, in the weights definitions; the convergence of the estimator of $Df^2_{\vepsilon}$, and in which context it is adequately approximated; and a more generic derivative-based generalization bound. In addition, VBSW demonstrated intriguing behaviors, like its impressive synergy with TCL \citep{transfer-curriculum-learning}, which would deserve more attention.

\section{Conclusion}

This work is based on the observation that, in supervised learning, a function $f$ is more difficult to approximate by a neural network in the regions where it is steep. We mathematically traduced this intuition, derived a generalization bound to illustrate it, and a methodology, Taylor Based Sampling, to test it empirically. In order to be able to use these insights for machine learning problems where $f$ is not available, we constructed a weighting scheme, Variance Based Samples Weighting (VBSW) that uses the variance of the training samples' labels to weight the training data set. VBSW is simple to use and implement because it only requires computing statistics on the input space. In Deep Learning, applying VBSW on the data set projected in an already trained neural network feature space allows reducing its error by simply re-training its last layer. Although specifically investigated in deep learning, this method applies to any loss-function-based supervised learning problem and is scalable, cost-effective, robust, and versatile. It is validated on several applications, such as glue benchmark with bert for text classification and regression, and Cifar10 with ResNet for image classification.











\newpage
\appendix

\section{Appendix A: Proofs}\label{appA}

\subsection{Illustration of the link using derivatives }
(Section \ref{sec:gb})\\

We look at approximating $f:\vx \rightarrow f(\vx)$, $\vx \in \mathbb{R}^{n_i}$, $f(\vx) \in \mathbb{R}^{n_o}$ with a NN $f_{\vtheta}$. The goal of the approximation problem can be seen as being able to generalize to points not seen during the training. We thus want the generalization error $\mathcal{J}_{\rvx}(\vtheta)$ to be as small as possible. Given an initial data set $\{\vx_1, ... , \vx_N\}$ drawn from $\rvx \sim d\mathbb{P}_{\rvx}$ and $\{f(\vx_1), ... , f(\vx_N)\}$, and the loss function $L$ being the squared $L_2$ error, recall that the integrated error $ J_{\rvx}(\vtheta) $, its estimation $\widehat{J_{\rvx}}(\vtheta)$ and the generalization error $\mathcal{J}_{\rvx}(\vtheta)$ can be written:

\begin{equation}\label{Jg}
\begin{split}
    J_{\rvx}(\vtheta) &= \int_{\mathbf{S}} \|f(\vx) - f_{\vtheta}(\vx)\|d\mathbb{P}_{\rvx},\\
    \widehat{J_{\rvx}}(\vtheta) &= \frac{1}{N} \sum_{i=1}^N \|f_{\vtheta}(\vx_i) - f(\vx_i)\big\|,\\
    \mathcal{J}_{\rvx}(\vtheta)&= J_{\rvx}(\vtheta) - \widehat{J_{\rvx}}(\vtheta),\\
\end{split}
\end{equation}

where $\|.\|$ denotes the squared $L_2$ norm. In the following, we find an upper bound for $ \mathcal{J}_{\rvx}(\vtheta) $. We start by finding an upper bound for $J_{\rvx}(\vtheta)$ and then for $ \mathcal{J}_{\rvx}(\vtheta) $ using \eqref{Jg}.

Let $ S_i$, $i \in \{1,...,N\}$ be some sub-spaces of a bounded space $\mathbf{S}$ such that $\mathbf{S} = \bigcup_{i=1}^N S_i$, $ \bigcap_{i=1}^N S_i =$ \O, and $\vx_i \in S_i$. Then,

\begin{equation*}
    \begin{split}
        J_{\rvx}(\vtheta) = &\sum_{i=1}^N \int_{S_i} \|f(\vx) - f_{\vtheta}(\vx)\|d\mathbb{P}_{\rvx}, \\
        J_{\rvx}(\vtheta) = &\sum_{i=1}^N \int_{S_i} \|f(\vx_i + \vx - \vx_i) - f_{\vtheta}(\vx) \|d\mathbb{P}_{\rvx}.\\
    \end{split}
\end{equation*}

Suppose that $n_i=n_o=1$ ($\vx$ becomes $x$ and $\rvx$ becomes $\rx$) and $f$ twice differentiable. Let $|\mathbf{S}| = \int_{\mathbf{S}} d \mathbb{P}_{\rx}$. The volume $|\mathbf{S}| = 1$ since $d \mathbb{P}_{\rx}$ is a probability measure, and therefore $|S_i| < 1$ for all $i \in \{1,...,N\}$ . Using Taylor expansion at order 2, and since $|S_i| < 1$ for all $i \in \{1,...,N\}$

\begin{equation*}
        J_{\rx}(\vtheta) = \sum_{i=1}^N \int_{S_i} \|f(x_i) + f'(x_i)(x - x_i) + \frac{1}{2}f''(x_i)(x - x_i)^2 - f_{\vtheta}(x)  + \mathcal{O}((x - x_i)^3)\|d\mathbb{P}_{\rx} .\\
\end{equation*}

To find an upper bound for $J(\vtheta)$, we can first find an upper bound for $|A_i(x)|$, with $A_i(x) = f(x_i) + f'(x_i)(x - x_i)  + \frac{1}{2}f''(x_i)(x - x_i)^2- f_{\vtheta}(x)  + \mathcal{O}((x - x_i)^3)$. 

NN $f_{\vtheta}$ is $K_{\vtheta}-$Lipschitz, so since $\mathbf{S}$ is bounded (so are $S_i$), for all $x \in S_i$, $|f_{\vtheta}(x) - f_{\vtheta}(x_i)| \leq K_{\vtheta}|x - x_i| $. Hence, 

\begin{equation*}
\begin{split}
    &f_{\vtheta}(x_i) - K_{\vtheta}|x - x_i| \leq f_{\vtheta}(x) \leq f_{\vtheta}(x_i) + K_{\vtheta}|x - x_i|,\\
    & -f_{\vtheta}(x_i) - K_{\vtheta}|x - x_i| \leq -f_{\vtheta}(x) \leq -f_{\vtheta}(x_i) + K_{\vtheta}|x - x_i|,\\
    & f(x_i) + f'(x_i)(x - x_i) + \frac{1}{2}f''(x_i(x - x_i)^2)  -f_{\vtheta}(x_i) - K_{\vtheta}|x - x_i|  + \mathcal{O}((x - x_i)^3) \\
    & \leq A_i(x) \leq f(x_i) + f'(x_i)(x - x_i)  + \frac{1}{2}f''(x_i)(x - x_i)^2 -f_{\vtheta}(x_i) + K_{\vtheta}|x - x_i|  + \mathcal{O}((x - x_i)^3),\\
    & A_i(x) \leq f(x_i)-f_{\vtheta}(x_i) + f'(x_i)(x - x_i)  + \frac{1}{2}f''(x_i)(x - x_i)^2  + K_{\vtheta}|x - x_i|  + \mathcal{O}((x - x_i)^3).
\end{split}
\end{equation*}

And finally, using triangular inequality,

\begin{equation*}
    \boxed{ A_i(x) \leq |f(x_i) - f_{\vtheta}(x_i)| + |f'(x_i)||x - x_i|  + \frac{1}{2}|f''(x_i)||x - x_i|^2   + K_{\vtheta}|x - x_i| +  \mathcal{O}(|x - x_i|^3).}
\end{equation*}

Now, $\|.\|$ being the squared $L_2$ norm: 
\begin{equation*}
\begin{split}
        J_{\rx}(\vtheta) = \sum_{i=1}^N \int_{S_i} \|&f(x_i) + f'(x_i)(x - x_i) + \frac{1}{2}f''(x_i)(x - x_i)^2- f_{\vtheta}(x)  + \mathcal{O}(|x - x_i|^3)\|d\mathbb{P}_{\rx}, \\
        J_{\rx}(\vtheta) \leq \sum_{i=1}^N \int_{S_i} &\Bigg[\Big(|f(x_i) - f_{\vtheta}(x_i)|\Big) + \Big(|f'(x_i)||x - x_i|  + \frac{1}{2}|f''(x_i)||x - x_i|^2   + K_{\vtheta}|x - x_i|\Big) \\
        &+  \mathcal{O}(|x - x_i|^3)\Bigg]^2d\mathbb{P}_{\rx},\\
        = \sum_{i=1}^N \int_{S_i} &\Bigg[|f(x_i) - f_{\vtheta}(x_i)|^2 \\
        &+ 2|f(x_i) - f_{\vtheta}(x_i)|\Big( |f'(x_i)||x - x_i| + \frac{1}{2}|f''(x_i)||x - x_i|^2  + K_{\vtheta}|x - x_i|\Big) \\
        &+ \Big[ \Big(|f'(x_i)||x - x_i| \Big)+ \Big(\frac{1}{2}|f''(x_i)||x - x_i|^2  + K_{\vtheta}|x - x_i|\Big)\Big]^2 +   \mathcal{O}(|x - x_i|^3)\Bigg]d\mathbb{P}_{\rx},\\
        = \sum_{i=1}^N \int_{S_i} &\Bigg[|f(x_i) - f_{\vtheta}(x_i)|^2 \\
        &+ 2|f(x_i) - f_{\vtheta}(x_i)|\Big( |f'(x_i)||x - x_i| + \frac{1}{2}|f''(x_i)||x - x_i|^2  + K_{\vtheta}|x - x_i|\Big) \\
          &+ \Big[ |f'(x_i)|^2|x - x_i|^2 + 2K_{\vtheta}|f'(x_i)||x - x_i|^2 +  K_{\vtheta}^2|x - x_i|^2\Big] +   \mathcal{O}(|x - x_i|^3)\Bigg]d\mathbb{P}_{\rx},\\
        = \sum_{i=1}^N \int_{S_i} &\Bigg[|f(x_i) - f_{\vtheta}(x_i)|^2 \\
        &+ 2|f(x_i) - f_{\vtheta}(x_i)|\Big( |f'(x_i)||x - x_i| + \frac{1}{2}|f''(x_i)||x - x_i|^2  + K_{\vtheta}|x - x_i|\Big) \\         
        &+ \Big( |f'(x_i)| +  K_{\vtheta}\Big)^2|x - x_i|^2 +   \mathcal{O}(|x - x_i|^3)\Bigg]d\mathbb{P}_{\rx}.
\end{split}
\end{equation*}

Hornik's theorem \cite{hornik} states that given a norm $\|.\|_{p, \mu} = $ such that $\|f\|^p_{p, \mu} = \int_{\mathbf{S}} |f(x)|^p d\mu(x)$, with $d\mu$ a probability measure, for any $\epsilon$, there exists $\vtheta$ such that for a Multi Layer Perceptron, $f_{\vtheta}$, $\| f(x) - f_{\vtheta}(x) \|^p_{p, \mu} < \epsilon$, 

This theorem grants that for any $\epsilon$, with $d\mu = \sum_{i=1}^{N} \frac{1}{N}\delta(x - x_i)$, there exists $\vtheta$ such that  

\begin{equation}\label{eq:hornik}
\begin{dcases}
    \| f(x) - f_{\vtheta}(x) \|^1_{1, \mu}  = \sum_{i=1}^{N}\frac{1}{N}|f(x_i) - f_{\vtheta}(x_i)| \leq \epsilon, \\
    \| f(x) - f_{\vtheta}(x) \|^2_{2, \mu}  = \sum_{i=1}^{N}\frac{1}{N}\big(f(x_i) - f_{\vtheta}(x_i)\big)^2 \leq \epsilon.  \\
\end{dcases}
\end{equation}

Let's introduce $i^*$ such that $i^* = \operatorname{argmin} |S_i|$. Note that for any $i \in \{1,...,N\}$, $\mathcal{O}(|S_i^*|^4)$ is $\mathcal{O}(|S_i|^4)$. Now, let's choose $\epsilon$ such that $\epsilon $ is $\mathcal{O}(|S_i^*|^4) $. Then, \eqref{eq:hornik} implies that 

\begin{equation*}
\begin{dcases}
    |f(x_i) - f_{\vtheta}(x_i)| = \mathcal{O}(|S_i|^4), \\
    \big(f(x_i) - f_{\vtheta}(x_i)\big)^2 =  \mathcal{O}(|S_i|^4), \\
    \widehat{J_{\rx}}(\vtheta) = \| f(x) - f_{\vtheta}(x) \|^2_{2, \mu} =  \mathcal{O}(|S_i|^4).\\
    \end{dcases}
\end{equation*}

Thus, we have $\mathcal{J}_{\rx}(\vtheta) = J_{\rx}(\vtheta) - \widehat{J_{\rx}}(\vtheta) = J_{\rx}(\vtheta)  + \mathcal{O}(|S_i|^4)$ and therefore,

\begin{equation*}
    \mathcal{J}_{\rx}(\vtheta) \leq \sum_{i=1}^N \int_{S_i}   \Big[\Big(|f'(x_i)| +  K_{\vtheta}\Big)^2|x - x_i|^2d\mathbb{P}_{\rx}\Big] +    \mathcal{O}(|S_i|^4).
\end{equation*}

Finally,

\begin{equation}\label{e1}
    \boxed{\mathcal{J}_{\rx}(\vtheta) \leq \sum_{i=1}^N   (|f'(x_i)| +  K_{\vtheta})^2\frac{|S_i|^3}{3} +    \mathcal{O}(|S_i|^4).}
\end{equation}

We see that on the regions where $f'(x_i) + K_{\vtheta}$ is higher, quantity $|S_i|$ (the volume of $S_i$) has a stronger impact on the GB. Then, since $|S_i|$ can be seen as a metric for the local density of the data set (the smaller $|S_i|$ is, the denser the data set is), the Generalization Bound (GB) can be reduced more efficiently by adding more points around $x_i$ in these regions. This bound also involves $K_{\vtheta}$, the Lipschitz constant of the NN, which has the same impact as $f'(x_i)$. It also illustrates the link between the Lipschitz constant and the generalization error, which has been pointed out by several works like, for instance, \cite{Gouk}, \cite{barlett} and \cite{qian2018lnonexpansive}.

\subsection{Problem 1: Unavailability of derivatives}
(Section \ref{sec:problem1})\\

In this paragraph, we consider $n_i >1$ but $n_o= 1$. The following derivations can be extended to $n_o > 1$ by applying it to $f$ element-wise. Let $\mathbf{e}\sim \mathcal{N}(0, \epsilon \mI_{n_i})$ with $\epsilon \in \mathbb{R}^+$, $\rve= (\epsilon_1,...,\epsilon_{n_i})$, i.e. $\epsilon_i \sim \mathcal{N}(0,\epsilon)$ and $\vepsilon = \epsilon(1,...,1)$. Using Taylor expansion on $f$ at order $2$ gives

\begin{equation*}
        f(\vx + \rve) = f(\vx) + \nabla_{\vx}f(\vx)\cdot \rve  + \frac{1}{2}\rve^T\cdot \mathbb{H}_x f(\vx) \cdot \rve + \mathcal{O}(\|\rve\|^3_2),
\end{equation*}

with $\nabla_x f$ and  $\mathbb{H}_x f(\vx)$ the gradient and the Hessian of $f$ w.r.t. $\vx$. We now compute $Var(f(X + \rve))$ and make $Df_{\vepsilon}^2(\vx) =   \epsilon \|\nabla_x f(\vx)\|^2_F + \frac{1}{2}\epsilon^2 \|\mathbb{H}_{\vx} f(\vx)\|^2_F$ appear in its expression to establish a link between these two quantities:

\begin{equation*}
\begin{split}
    Var(f(\vx + \rve)) &= Var\Big(f(\vx) + \nabla_{\vx}f(\vx)\cdot \rve  + \frac{1}{2}\rve^T\cdot \mathbb{H}_x f(\vx) \cdot \rve + \mathcal{O}(\|\rve\|^3_2)\Big), \\
    &= Var\Big(\nabla_{\vx}f(\vx)\cdot \rve  + \frac{1}{2}\rve^T\cdot \mathbb{H}_x f(\vx) \cdot \rve\Big) +\mathcal{O}(\|\vepsilon\|^3_2). 
\end{split}
\end{equation*}

Since $\epsilon_i \sim \mathcal{N}(0,\epsilon)$,  $\vx = (x_1,...,x_{n_i})$ and with $\frac{\partial^2f}{\partial x_i x_j}(\vx)$ the cross derivatives of $f$ w.r.t. $x_i$ and $x_j$,

\begin{equation*}
\begin{split}
    \nabla_{\vx}f(\vx)\cdot \rve   + \frac{1}{2}\rve^T\cdot \mathbb{H}_x f(\vx) \cdot \rve  =& \sum_{i = 1}^{n_i}  \epsilon_i \frac{\partial f}{\partial x_i}(\vx) +\frac{1}{2} \sum_{j = 1}^{n_i} \sum_{k = 1}^{n_i} \epsilon_j \epsilon_k \frac{\partial^2f}{\partial x_j x_k}(\vx), \\
\end{split}
\end{equation*}
\begin{equation*}
\begin{split}
    Var\Big(\nabla_{\vx}f(\vx)\cdot \rve   + \frac{1}{2}\rve^T\cdot \mathbb{H}_x f(\vx) \cdot \rve\Big)  = &Var \Big(\sum_{i = 1}^{n_i}  \epsilon_i \frac{\partial f}{\partial x_i}(\vx) + \frac{1}{2}\sum_{j = 1}^{n_i} \sum_{k = 1}^{n_i} \epsilon_j \epsilon_k \frac{\partial^2f}{\partial x_j x_k}(\vx)\Big),\\
    =& \sum_{i_1 = 1}^{n_i} \sum_{i_2 = 1}^{n_i} Cov \Big(\epsilon_{i_1} \frac{\partial f}{\partial x_{i_1}}(\vx), \epsilon_{i_2} \frac{\partial f}{\partial x_{i_2}}(\vx) \Big),\\
    &+ \frac{1}{4}\sum_{j_1 = 1}^{n_i} \sum_{k_1 = 1}^{n_i}  \sum_{j_2 = 1}^{n_i} \sum_{k_2 = 1}^{n_i} Cov \Big(\epsilon_{j_1} \epsilon_{k_1} \frac{\partial^2f}{\partial x_{j_1} x_{k_1}}(\vx), \epsilon_{j_2} \epsilon_{k_2} \frac{\partial^2f}{\partial x_{j_2} x_{k_2}}(\vx)\Big)\\
    &+ \sum_{i = 1}^{n_i} \sum_{j = 1}^{n_i} \sum_{k = 1}^{n_i} Cov \Big(\epsilon_{i} \frac{\partial f}{\partial x_i}(\vx), \epsilon_j \epsilon_k \frac{\partial^2f}{\partial x_j x_k}(\vx)\Big),\\
    =& \sum_{i_1 = 1}^{n_i} \sum_{i_2 = 1}^{n_i} \frac{\partial f}{\partial x_{i_1}}(\vx)\frac{\partial f}{\partial x_{i_2}}(\vx) Cov \Big(\epsilon_{i_1} , \epsilon_{i_2}  \Big)\\
    &+ \frac{1}{4}\sum_{j_1 = 1}^{n_i} \sum_{k_1 = 1}^{n_i}  \sum_{j_2 = 1}^{n_i} \sum_{k_2 = 1}^{n_i} \frac{\partial^2f}{\partial x_{j_1} x_{k_1}}(\vx)  \frac{\partial^2f}{\partial x_{j_2} x_{k_2}}(\vx)Cov \Big(\epsilon_{j_1} \epsilon_{k_1} , \epsilon_{j_2} \epsilon_{k_2}\Big)\\
    &+ \sum_{i = 1}^{n_i} \sum_{j = 1}^{n_i} \sum_{k = 1}^{n_i} \frac{\partial f}{\partial x_i}(\vx)\frac{\partial^2f}{\partial x_j x_k}(\vx)Cov \Big(\epsilon_{i} , \epsilon_j \epsilon_k \Big).\\
\end{split}
\end{equation*}

In this expression, three quantities have to be assessed : $Cov (\epsilon_{i_1} ,\epsilon_{i_2} )$, $Cov(\epsilon_i, \epsilon_j\epsilon_k)$ and $Cov(\epsilon_{j_1}\epsilon_{k_1}, \epsilon_{j_2}\epsilon_{k_2})$.\\

First, since $(\epsilon_1,...,\epsilon_{n_i})$ are i.i.d., 

\begin{equation*}
Cov \Big(\epsilon_{i_1} ,\epsilon_{i_2} \Big) = 
\begin{dcases}
Var(\epsilon_i) = \epsilon \text{ if } i_1=i_2=i, \\
0 \text{ otherwise.}
\end{dcases}.
\end{equation*}

To assess $Cov(\epsilon_i, \epsilon_j\epsilon_k)$, three cases have to be considered.
\begin{itemize}
    \item If $i = j = k$, because $\mathbb{E}[\epsilon_i^3] = 0$,
        \begin{equation*}
            \begin{split}
                Cov(\epsilon_i, \epsilon_j\epsilon_k) &= Cov(\epsilon_i, \epsilon_i^2),\\
                &= \mathbb{E}[\epsilon_i^3] - \mathbb{E}[\epsilon_i]\mathbb{E}[\epsilon_i^2],\\
                &= 0.
            \end{split}
        \end{equation*}
    \item If $i = j$ or $i = k$ (we consider $i=k$, and the result holds for $i=j$ by commutativity),
        \begin{equation*}
            \begin{split}
                Cov(\epsilon_i, \epsilon_j\epsilon_k) &= Cov(\epsilon_i, \epsilon_i\epsilon_j),\\
                &= \mathbb{E}[\epsilon_i^2 \epsilon_j] - \mathbb{E}[\epsilon_i]\mathbb{E}[\epsilon_i\epsilon_j],\\
                &= \mathbb{E}[\epsilon_i^2 ]\mathbb{E}[\epsilon_j],\\
                &= 0.
            \end{split}
        \end{equation*}
    \item If $i \neq j$ and $i \neq k$, $\epsilon_i$ and $\epsilon_j\epsilon_k$ are independent and so $Cov(\epsilon_i, \epsilon_j\epsilon_k)$ = 0.
\end{itemize}
Finally, to assess  $Cov(\epsilon_{j_1}\epsilon_{k_1}, \epsilon_{j_2}\epsilon_{k_2})$, four cases have to be considered:
\begin{itemize}
    \item If $j_1 = j_2 = k_1 = k_2 = i$,
    \begin{equation*}
        \begin{split}
            Cov(\epsilon_{j_1}\epsilon_{k_1}, \epsilon_{j_2}\epsilon_{k_2}) &= Var(\epsilon_i^2),\\
            &= 2\epsilon^2.
        \end{split}
    \end{equation*}
    \item If $j_1 = k_1 = i$ and $j_2 = k_2 = j$, $Cov(\epsilon_{j_1}\epsilon_{k_1}, \epsilon_{j_2}\epsilon_{k_2}) = Cov(\epsilon_i^2, \epsilon_j^2) = 0 $ since $\epsilon_i^2$ and $\epsilon_j^2$ are independent.
    \item If $j_1 = j_2 = j$ and $k_1 = k_2 = k$,
    \begin{equation*}
        \begin{split}
            Cov(\epsilon_{j_1}\epsilon_{k_1}, \epsilon_{j_2}\epsilon_{k_2}) &= Var(\epsilon_{j}\epsilon_{k}),\\
            & = Var(\epsilon_{j}) Var(\epsilon_{k}),\\
            &= \epsilon^2.
        \end{split}
    \end{equation*}
    \item If $j_1 \neq k_1, j_2$ and $k_2$,
    \begin{equation*}
        \begin{split}
            Cov(\epsilon_{j_1}\epsilon_{k_1}, \epsilon_{j_2}\epsilon_{k_2}) &= \mathbb{E}[\epsilon_{j_1}\epsilon_{k_1} \epsilon_{j_2}\epsilon_{k_2}] - \mathbb{E}[\epsilon_{j_1}\epsilon_{k_1}]\mathbb{E}[\epsilon_{j_2}\epsilon_{k_2}],\\
            &= \mathbb{E}[\epsilon_{j_1}]\mathbb{E}[\epsilon_{k_1} \epsilon_{j_2}\epsilon_{k_2}] - \mathbb{E}[\epsilon_{j_1}]\mathbb{E}[\epsilon_{k_1}]\mathbb{E}[\epsilon_{j_2}\epsilon_{k_2}],\\
            &= 0.
        \end{split}
    \end{equation*}
\end{itemize}
All other possible cases can be assessed using the previous results, commutativity and symmetry of $Cov$ operator.
Hence,
\begin{equation*}
\begin{split}
    Var\Big(\nabla_{\vx}f(\vx)\cdot \rve   + \frac{1}{2}\rve^T\cdot \mathbb{H}_x f(\vx) \cdot \rve\Big)  =& \sum_{i_1 = 1}^{n_i} \sum_{i_2 = 1}^{n_i} \frac{\partial f}{\partial x_{i_1}}(\vx)\frac{\partial f}{\partial x_{i_2}}(\vx) Cov \Big(\epsilon_{i_1} , \epsilon_{i_2}  \Big)\\
    &+ \frac{1}{4}\sum_{j_1 = 1}^{n_i} \sum_{k_1 = 1}^{n_i}  \sum_{j_2 = 1}^{n_i} \sum_{k_2 = 1}^{n_i} \frac{\partial^2f}{\partial x_{j_1} x_{k_1}}(\vx)  \frac{\partial^2f}{\partial x_{j_2} x_{k_2}}(\vx)Cov \Big(\epsilon_{j_1} \epsilon_{k_1} , \epsilon_{j_2} \epsilon_{k_2}\Big),\\
    = &\sum_{i=1}^{n_i} \epsilon\frac{\partial f^2}{\partial x_i}(\vx)  + \frac{1}{2}\sum_{j = 1}^{n_i} \sum_{k = 1}^{n_i} \epsilon^2 \frac{\partial^2f^2}{\partial x_j x_k}(\vx),\\
    = & \epsilon \|\nabla_x f(\vx)\|^2_F + \frac{1}{2}\epsilon^2 \|\mathbb{H}_{\vx} f(\vx)\|^2_F,\\
    = & Df_{\vepsilon}^2(\vx).
\end{split}
\end{equation*}

And finally, 

\begin{equation}\label{e2}
    \boxed{Var(f(\vx + \rve)) = Df_{\vepsilon}^2(\vx) +\mathcal{O}(\|\vepsilon\|^3_2)}
\end{equation}

If we consider $\widehat{Df^2}_{\epsilon}(\vx)$ as defined in \eqref{dn}, on section* ~\ref{sec:tbs} of the main document, $\widehat{Df^2}_{\epsilon}(\vx) \underset{k \rightarrow \infty}{\rightarrow} Var(f(\vx + \vepsilon) )$ . Since $Var(f(\vx + \vepsilon) ) = Df^2_{\vepsilon}(\vx) + \mathcal{O}(\|\vepsilon\|^3_2)$, $\widehat{Df^2}_{\epsilon}(\vx)$ is a biased estimator of $Df^2_{\vepsilon}(\vx)$, with bias $\mathcal{O}(\|\vepsilon\|^3_2)$. Hence, when $\epsilon \rightarrow 0$, $\widehat{Df^2}_{\epsilon}(\vx)$ becomes an unbiased estimator of $Df^2_{\vepsilon}(\vx)$. 

\newpage

\section{Appendix B: Hyperparameters spaces}\label{appB}

The values chosen for the hyperparameters experiments are gathered in Table \ref{hyp-table}. For Adam optimizer hyperparameters, we kept the default values of Keras implementation. We chose these hyperparameters after simple grid searches.\\

\begin{table}[h]
  \centering
  \begin{tabular}{llllllll}
    Experiment & $m$ & $k$ & learning rate & batch size & epochs & optimizer & random seeds\\
      \hline
    double moon & 100 & 20 & $1 \times 10^{-3}$ & 100  & 10000 & SGD & 50\\
    Boston housing & 8 & 35 & $5 \times 10^{-4}$ & 404  & 50000  & Adam & 10\\
    Breast Cancer & 50 & 35 & $5 \times 10^{-2}$ & 455  & 250000  & Adam & 10\\
    MNIST & 40 & 20 & $1 \times 10^{-3}$  & 25  & 25 & Adam& 40\\
    Cifar10 & 40 & 20 & $1 \times 10^{-3}$ & 25  & 25  &Adam & 50\\
    RTE & 20 & 10 & $3 \times 10^{-4}$ & 8  & 10000  & Adam & 50\\
    STS-B & 30 & 30 & $3 \times 10^{-4}$ & 8  & 10000  & Adam & 50\\
    MRPC & 75 & 25 & $3 \times 10^{-4}$ & 16  & 10000  & Adam & 50\\
  \end{tabular}
  \caption{\label{hyp-table} Hyperparameters values for experiments.}
\end{table}

\bibliographystyle{plain}

\bibliography{refs}
\end{document}